\documentclass[12pt]{article}

\usepackage{etoolbox}
\usepackage[top=1in,bottom=1in, left=1in, right=1in]{geometry}
\usepackage{mathtools}
\usepackage{enumitem}
\usepackage{amssymb}
\usepackage{graphicx}
\usepackage{amsmath}
\usepackage{xcolor}
\usepackage{relsize}
\usepackage{breqn}
\usepackage{amsmath,amsthm,amssymb,amsopn,amsfonts,pdfpages,algorithmic,algorithm,dsfont}
\usepackage{graphics}
\usepackage{graphicx}
\usepackage{subfig}
\usepackage{caption,url}
\usepackage{stackengine}
\usepackage[colorlinks=true,linkcolor=cyan,citecolor=blue]{hyperref}
\usepackage{authblk}

\newtheorem{theorem}{Theorem}[section]

\newtheorem{remark}{Remark}[section]

\newtheorem{defn}{Definition}[section]

\newtheorem{definition}{Definition}[section]

\newcommand{\bde}{\begin{defn}}
\newcommand{\ede}{\end{defn}}

\newcommand{\hg}{\mathcal{HG}}
\newcommand{\hgs}{\mathcal{HGS}}
\newcommand{\hgsm}{\mathcal{HGS}_m^{\vec m}}
\newcommand{\hgsn}{\mathcal{HGS}_n^{\vec n}}
\newcommand{\hgn}{\hg_n^{\vec n}}
\newcommand{\hgm}{\hg_m^{\vec m}}

\newcommand{\gn}{\mathcal{G}_n}

\newcommand{\bD}{\boldsymbol{\Delta}}

\newcommand{\calT}{\mathcal{T}}

\makeatletter
\newcommand{\distas}[1]{\mathbin{\overset{#1}{\kern\z@\sim}}}%
\newsavebox{\mybox}\newsavebox{\mysim}
\newcommand{\distras}[1]{%
  \savebox{\mybox}{\hbox{\kern3pt$\scriptstyle#1$\kern3pt}}%
  \savebox{\mysim}{\hbox{$\sim$}}%
  \mathbin{\overset{#1}{\kern\z@\resizebox{\wd\mybox}{\ht\mysim}{$\sim$}}}%
}
\makeatother

\title{Subgraph nomination: Query by Example Subgraph Retrieval in Networks}

\author{ Al-Fahad M.~Al-Qadhi$^1$, Carey E. Priebe$^2$,\\\vspace{-2mm}
Hayden S. Helm$^3$, Vince Lyzinski$^1$\\
\vspace{3mm}
\small{$^1$ University of Maryland, College Park, Department of Mathematics\\
$^2$ Johns Hopkins University, Department of Applied Mathematics and Statistics\\
$^3$ Johns Hopkins University, Center for Imaging Science}
}
\begin{document}

\maketitle

%

\begin{abstract}
This paper introduces the subgraph nomination inference task, in which example subgraphs of interest are used to query a network for similarly interesting subgraphs.
	This type of problem appears time and again in real world problems connected to, for example, user recommendation systems and structural retrieval tasks in social and biological/connectomic networks.
	We formally define the subgraph nomination framework with an emphasis on the notion of a user-in-the-loop in the subgraph nomination pipeline. In this setting, a user can provide additional post-nomination light supervision that can be incorporated into the retrieval task. After introducing and formalizing the retrieval task, we examine the nuanced effect that user-supervision can have on performance, both analytically and across real and simulated data examples.
	\end{abstract}

\section{Introduction}
Stated succinctly, the subgraph nomination problem is as follows:  
given a subgraph or subgraphs of interest in a network $G_1$, we seek to find heretofore unknown subgraphs of interest in $G_1$ or in a second network $G_2$.
The subgraph nomination problem can be viewed as an amalgam of the problems of noisy subgraph detection \cite{nsg1,nsg2,nsg3,bertozzi_2018,sussman2018matched} and vertex nomination \cite{marchette2011vertex,coppersmith2014vertex,FisLyzPaoChePri2015,lyzinski2017consistent}, in which our task is to produce a rank-list of candidate subgraphs, with (ideally) unknown subgraphs of interest concentrating at the top of the rank list.
While seemingly simple on its surface, subgraph nomination necessarily entails the often
non-trivial combination of 
subgraph detection (i.e., we have to find candidate subgraphs to rank) and multiple graph comparison methodologies  (i.e., we have to rank the candidate subgraphs).
We note here that the subgraph nomination problem considered herein is closely related to many existing graph query problem formulations (see, for example, the work on query by example in \cite{exemplar1,exemplar2}).  Our novel contribution is the novel statistical framework underlying our subgraph nomination formulation, which allows us to frame and rigorously study the problem (and the effect of a user-in-the-loop) in a number of popular statistical network models, with the ultimate goal of establishing results of statistical consistency for subgraph nomination akin to the analogues in the vertex nomination literature \cite{lyzinski2016consistency, lyzinski2017consistent}.

\emph{Subgraph detection} ---here, producing candidate subgraphs of interest from $G_1$ or $G_2$---is an active area of research in machine learning and pattern recognition, and encompasses both the famous subgraph isomorphism problem (see, for example, \cite{sg1,sg2,sg3,sg4}) as well as numerous noisy subgraph detection routines.
In subgraph nomination, subgraph detection is further complicated by the generality of the (topological) features that define the training subgraphs as ``interesting."
In particular, it may be the case that the query set encompasses multiple different interesting templates \cite{bertozzi_2018, nsg1} or motifs-of-interest \cite{bio3, lyzinski2015community}, each delineating an activity or structure of interest in the network. 
Moreover, the templates of interesting subgraphs need not be manifestly similar to each other.
The notion of \emph{similarity} across subgraphs will be formalized via a subgraph dissimilarity mapping $\boldsymbol{\Delta}$, and a subgraph nomination scheme can (loosely) be considered as an estimate of the subgraph structure of $g_2$ combined with a dissimilarity $\bD$ used to rank the subgraphs (see Def. \ref{def:SGN}).
If $\mathfrak{S}_{G}$ represents the collection of all subgraphs of a network $G$, then in the single-graph setting
$$\boldsymbol{\Delta}:\mathfrak{S}_{G_1}\times \mathfrak{S}_{G_1}\rightarrow[0,1],
$$
and in the pair of graphs setting
$$\boldsymbol{\Delta}:\mathfrak{S}_{G_1}\times \mathfrak{S}_{G_2}\rightarrow[0,1],
$$
with smaller values indicating more similar subgraphs.
We comment here that although the two graph setting will be our focus moving forward, this naturally lifts to the single graph setting by considering the two graphs to nominate across as parts of a partition of a larger network.

Large values of $\boldsymbol{\Delta}$ indicate that the subgraphs are highly dissimilar according to $\bD$, and
candidate subgraphs are defined as interesting if they are sufficiently similar (i.e., insufficiently dissimilar) to \emph{any} subgraph in the training sample.
Choosing an appropriate $\bD$ from which to construct the subgraph rankings is of primary import, and adaptive methods (similar to those in the learning-to-rank problem in the information retrieval literature \cite{liu2011learning,helm2020learning}) can be considered to learn $\bD$ from the training subgraphs of interest.
While estimating an optimal $\bD$ is central in this problem framework, we do not derive formal procedures for estimating $\bD$ in this paper. 
Instead, we choose to focus on the effect of user-in-the-loop supervision across a variety of possible $\bD$ choices, where the user is modeled as additional light supervision that can be used to refine the output of a subgraph nomination routine.

As a motivating examples for our focus on studying the effects of a user-in-the-loop, we consider the problem of detecting and nominating  particular brain regions of interest across the hemispheres of a human connectome from the BNU1 dataset \cite{zuo2014open} downloaded from \url{https://neurodata.io/mri/}, and finding subreddits across different time periods in the Reddit social network data collected in \cite{baumgartner2020pushshift} and downloaded from \url{https://files.pushshift.io/reddit/comments/}. 

In the setting of BNU1, subgraph nomination can be used as a tool to better understand (and detect) the structural similarity between brain regions across hemispheres.  To this end, we
consider a region (or regions) of interest in the left hemisphere, and there might be multiple subgraphs in the right hemisphere that match (according to $\boldsymbol{\Delta}$) well to the training data.
Moreover, the most similar region of interest in the right hemisphere (i.e., the ${\boldsymbol{\Delta}}$-optimal subgraph of the right hemisphere) may vary dramatically depending on the $\boldsymbol{\Delta}$ used (see, for example, Figure 7 in \cite{sussman2018matched}), and the optimal region may not significantly overlap with a true latent region of interest in the left hemisphere.
In this case, even a sensible ranking scheme would be potentially unable to correctly retrieve the desired region of interest in the right hemisphere (even if the proper subgraph was identified). In this case, we can naturally employ the help of a user-in-the-loop \cite{amershi2014power},
who, given a vertex, can decide whether it is interesting/part of an interesting subgraph to help refine our ranking.

Meanwhile, in the setting of the Reddit social network, subgraph nomination can be used to track subreddits across time based on their structure. 
Therefor, we consider a subreddit of interest in a week (based on posting time) and attempt to find it among unlabeled subreddits in the next week. For the $\boldsymbol{\Delta}$, we use the Graph Matching Network (GMN) from \cite{li2019graph}. 
Here, the graph matching network training starts quick, but once a certain level of accuracy is passed the training slows significantly and oscillates. 
We investigate adding the user in the loop to improve the accuracy with less training. 
Accuracy in this setting represents the chance of finding the subreddit among the top $l$ nominated graphs. 
As can be seen in \ref{fig:GMN+VN}, the user in the loop can achieve better accuracy with less training and at a lower level, even when the user is relatively faulty. 
For large networks, such as Reddit this could save computational resources, as well as human resources by reducing the amount of subreddits an expert needs to investigate.

The subgraph nomination framework we present in this paper is both general enough to include the operationally significant effect of a user-in-the-loop and principled enough to theoretically analyze nomination schemes. After discussing preliminaries and presenting the problem framework, we 
demonstrate multiple pathologies to highlight the power of the subsequent theory as a predictive tool of the efficacy of the user in general situations. In particular, we show that a user can improve the performance over a scheme that achieves the Bayes optimal rate sans user-in-the-loop, even in the presence of potential user-error. This result is similar to results in the classification literature when given noisy training labels \cite{frenay2013classification,natarajan2013learning}. We further show that, given noisily recovered subgraphs, including information from an oracle user can deteriorate performance due to the noise within the subgraph detection routines.

\vspace{3mm}

\noindent{\bf Notation:} For a positive integer $n$, we will denote $[n]:=\{1,2,3,\cdots,n\}$; $J_n$ to the $n\times n$ matrix of all $1$'s; $\mathcal{G}_n$ to be the set of labeled $n$-vertex graphs; for $G\in\mathcal{G}_n$, and $W\subset V(G)$, we let $G[W]$ denote the induced subgraph of $G$ on $W$.


\section{Introduction and Background}
\label{Motivation}

In subgraph nomination, our task is as follows: given a subgraph or subgraphs of interest in a network $G_1$, we seek to find heretofore unknown subgraphs of interest in a second network $G_2$.
Our approach to subgraph nomination proceeds by
\begin{itemize}
	\item[1.] Partitioning the vertices in $G_2$ into candidate subgraphs (note that the choice of non-overlapping candidate subgraphs is merely a theoretical/notational convenience, and in practice, the subgraphs to be ranked can have overlap);
	\item[2.] Ordering the candidate subgraphs in $G_2$ into a rank list with the unknown subgraphs of interest ideally concentrating at the top of the rank list.
\end{itemize}

Subgraph nomination is a natural extension of vertex nomination (VN) \cite{marchette2011vertex,coppersmith2014vertex,suwan2015bayesian,Fishkind_2015,lyzinski2017consistent,agterberg2019vertex}, and we will begin by providing a brief background on recent developments in VN, as this will provide useful context for our later novel formulation of subgraph nomination.

\subsection{Vertex Nomination}
\label{sec:vn}

Semi-supervised querying of large graphs and databases is a common graph inference task. 
For example, in a graph database with multiple features, one may be given a collection of book names all with the common \emph{latent} feature ``Horror Fiction Best Sellers,'' and the goal would be to query the database for more titles that fit this description; see, for example, the work in \cite{angles2016foundations,rastogi2017vertex}. 
Learning this unifying feature from the query set is a challenging problem unto itself \cite{rastogi2019neural}, especially in settings where the features that define interestingness are nuanced or multi-faceted.

While we can interpret such a problem as a vertex classification problem where vertices are either labeled interesting or not, in the presence of very large networks with large class imbalance between interesting and non-interesting vertices, there is often  limited training data and limited user resources for verifying returned results.
In this setting, an information retrieval (IR) framework may be more appropriate, wherein unlabeled vertices would be ordered in a rank list based on how interesting they are deemed to be.
This inference task is synonymous with \emph{vertex nomination} (or personal recommender systems on graphs).

In \cite{patsolic2017vertex,lyzinski2017consistent,agterberg2019vertex,levin2020role}, the vertex nomination problem is defined as follows.  
Given vertices of interest $V^*\subset V(G_1)$ in a network $G_1$, and corresponding unknown vertices of interest $U^*\subset V(G_2)$ in a second network $G_2$ (whose identities are hidden to the user), use $G_1,G_2,V^*$ to rank the vertices in $G_2$ into a nomination list, with vertices in $U^*$ concentrating at the top of the nomination list.
In its initial formulation \cite{coppersmith2014vertex,marchette2011vertex,Fishkind_2015,yoder2018vertex}, the feature that defined vertices as interesting was membership in a community of interest, and the community memberships of vertices in $V^*$ were used to nominate the vertices in $G_1$ (no second network was introduced, or $G_2=G_1\setminus\{V^*\}$) with unknown community memberships.
Subsequent work \cite{rastogi2017vertex,rastogi2019neural,patsolic2017vertex,lyzinski2017consistent} sought to generalize the features that defined vertices as interesting beyond simple community membership, and lifted the vertex nomination problem to the two graph setting.

In order to allow for a broadly general class of networks to be considered, the general vertex nomination problem framework of \cite{agterberg2019vertex,lyzinski2017consistent} referenced above is situated in the context of nominatable distributions, a broad class of random graph distributions defined in  \cite{agterberg2019vertex}.
Within this broad class of models, the concepts of Bayes optimality and consistency were developed in \cite{lyzinski2017consistent, agterberg2019vertex} and the important result that universally consistent VN schemes do not exist is proven in \cite{lyzinski2017consistent}.
These results were leveraged in \cite{agterberg2019vertex} to develop an probabilistic adversarial contamination model in the context of VN as well as regularization schemes to counter the adversary.
The results of \cite{lyzinski2017consistent, agterberg2019vertex} were further extended to the richly featured network setting in \cite{levin2020role}, in which the (potentially) complementary roles of features and network structure are explored in the VN task.

Beyond the development of novel VN algorithms \cite{yoder2018vertex}, one of the key aspects of the recent theoretical developments in VN is the notion that vertex labels in $g_2$ are uninformative in the ranking scheme, which is sensible if we are mirroring setting in which the vertex labels do not aid in the delineation between interesting and uninteresting vertices.
In subgraph nomination, the uninformative nature of the labels can be accounted for in the dissimilarity $\bD$ (see Eq. \ref{eq:D}).  Note that while similar consistency results to those in VN can be derived in the setting of subgraph nomination, we do not focus on that here.
Rather we will focus on the role of users-in-the-loop in the subgraph nomination framework.

\begin{remark}
We note that it is possible to view the subgraph nomination methods as naturally being built upon existing VN methods via uncovering vertices of interest (from VN) and building out from them to uncover subgraph structure of interest.  One immediate hurdle to this approach would be if the subgraph structure is the interesting trait rather than the individual vertex identities, as then the nominated vertices in VN would need to look at local structure to be correctly retrieved, and essentially subgraphs would end up being the nominated structures.  It is also natural to use subgraph nomination for VN, by nominating structures first and then vertices within the structures.
\end{remark}

\subsection{Hierarchical Subgraph Models}
\label{sec:HG}

We will use $\gn$ to denote the set of $n$-vertex labeled graphs on a common vertex set $V=V_n$.
Given the (informal) explanation of the motivating task in subgraph nomination---given light supervision, partition a graph into subgraphs and rank the subgraphs based on how interesting they are judged to be---hierarchical network models are a natural setting for initially formalizing the subgraph nomination inference task.
We note here that the subgraph nomination inference task can be formulated in the context of any graph division mechanism that yields partition subgraph structure (e.g., graph cuts \cite{cuts,cuts2}, graph clustering methods \cite{newman2006modularity,blondel2008fast,traag2019louvain}, etc.); the Hierarchical blockmodel considered below is one such setting, and it also provides a convenient model setting in which to formulate subgraph nomination formally.

Hierarchical models have seen a surge in popularity in the network literature (see, for example,
\cite{sales-pardo07:_extrac,clauset08:_hierar,park10:_dynam_bayes,Peixoto_HSBM,lyzinski2015community,bickel_hierarchical}) and naturally allow for multiple layers of structure to be simultaneously modeled in a network, allowing us to imbue a graph $G$ with subgraph structures of a desired form or motif-type.
We begin with our definition of a hierarchical network as a network together with a \emph{Network Hierarchical Function}.
\begin{figure*}[t!]
	\centering
	\includegraphics[width=1\textwidth]{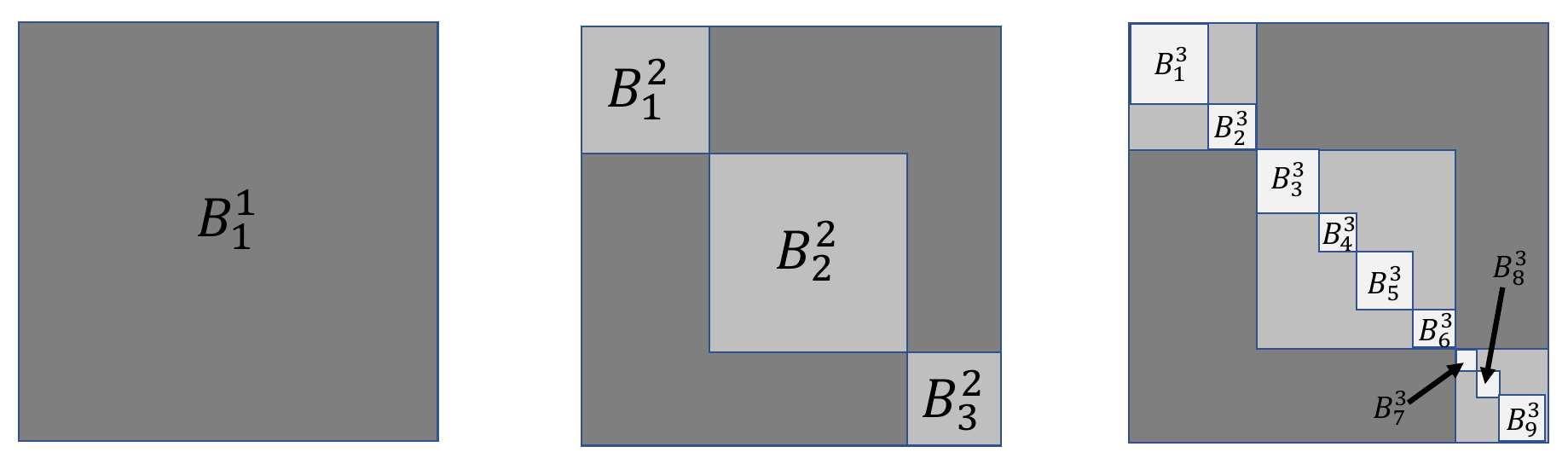}
	\caption{An example of a 3-Level hierarchical network (where the three levels are shown in the right panel, the top two levels in the middle panel, and the highest level in the left panel).
		The partition provided by $H(\cdot,1)$ is in darkest grey; the partition provided by $H(\cdot,2)$ is in medium grey; and the partition provided by $H(\cdot,3)$ is in lightest grey.}
	\label{fig:hier_ex}
\end{figure*}
\theoremstyle{definition}
\begin{definition}
	Let $g=(V,E)\in\gn$, and for each $i\in\mathbb{Z}>0$ let $[i]=\{1,2,\cdots,i\}$.
	Let $k\leq n$.
	A function 
	$$H=H_k:V\times \big [k]\longrightarrow \big [n]$$ 
	is a \emph{k-level Network Hierarchical Function of [n]} if 
	\begin{itemize}
		\item[i.] For each $i\in[k]$, $H(\cdot,i)$ represents a partition of the vertices of $V$ into $n_i:=n_i^{(H)}$ nonempty parts (so that $\max_v H(v,i)=n_i$).
		These $n_i$ satisfy
		$1\leq n_1\leq n_2\leq\cdots\leq n_k\leq n.$
		The \emph{signature} of $H$ is defined to be the vector $\vec n=(n_1,n_2,\cdots,n_k)$.
		Note that by considering appending $0$'s onto the end of $\vec n$ as needed to make it length $n$, we can consider the signature of an $n$ vertex hierarchical graph to be a length $n$ vector.
		\item[ii.] $H(v_1,j)=H(v_2,j)$ only if $H(v_1,i)=H(v_2,i)$ for all $i\leq j$; i.e., the partition is nested.
	\end{itemize}
	For an example of network hierarchical functions, see Figure \ref{fig:hier_ex}.
	A graph $g\in\gn$ together with a Network Hierarchical Function $H_k$ of $[n]$ is a \emph{k-level Hierarchical Graph}. 
	For $k\leq n$, we denote 
	
	\begin{align*}
	    \mathcal{H}_{k,n}=\{H_k\,|\,H_k\text{ is a }&\text{k-level Network}\\
	    & \text{Hierarchical Function of }[n]\},
	\end{align*}
	and we define
	$$\hg_n:=\{(g,H)\,|\,g\in\gn,\text{ and }H\in \cup_{k=1}^n\mathcal{H}_{k,n}\}$$ 
	to be the set of all Hierarchical graphs of order $n$.
\end{definition}
\noindent
Letting $k\leq n$, let $H\in\mathcal{H}_{k,n}$. 
For each $i\in [k]$ and $j\in[n_i]$, we define the sets 
$$B^i_j=\{v\in V(g)|H(v,i)=j\}$$ to be the set of vertices in the 
$j$-th part of the $i$-th level of the hierarchy;
\begin{align*}
    B_{(j)}^{i-1}=\{v\in V(G)\,&\text{ s.t. for all }v'\in B_j^i,\\
    & H(v,i-1)=H(v',i-1)\}
\end{align*}
to be the one-step upward merge of $B^i_j$ in the hierarchy;
$$B^i=\{B^i_j\,|\, 1<j<n_i\}$$ 
to be the set of all parts at level $i$ in the hierarchy; and $$B=\{B^i:i\in[k]\}.$$ 
to be the set of blocks.

As an example, consider the hierarchical network in Figure \ref{fig:hier_ex}, which shows a 3-level hierarchical network, where we have (for example):
\begin{align*}
	n_1&=1; n_2=3; n_3=9;\\
	B_1^1&=V=B_2^1\cup B_2^2\cup B_2^3;\\
	B^2_{(3)}&=B^3_3\cup B^3_4\cup B^3_5\cup B^3_6 ;\\
	B^3&=\{B^3_1,B^3_2,B^3_3,B^3_4,B^3_5,B^3_6,B^3_7,B^3_8,B^3_9 \};\\
	B&=\{B^1_1,B^2_1,B^2_2,B^2_3,B^3_1,B^3_2,\\
	&~~~~~~B^3_3,B^3_4,B^3_5,B^3_6,B^3_7,B^3_8,B^3_9\}.
\end{align*}

\subsubsection{Hierarchical Stochastic Blockmodels}
\label{sec:HSBM}

One important example of a network distribution with hierarchical structure  is the hierarchical stochastic blockmodel (HSBM) \cite{Peixoto_HSBM,lyzinski2015community}.
Before defining the HSBM, we first define the standard stochastic blockmodel \cite{sbm}.
\begin{definition}
	We say that an $n$-vertex random graph $G$ is an instantiation of a stochastic blockmodel with parameters $(n,K,\Lambda,\pi)$ 
	(abbreviated $A\sim\text{SBM}(n,K,\Lambda,\pi)$) if 
	\begin{itemize}
		\item[i.] The block membership vector $\pi\in\mathbb{R}^K$ satisfies $\pi(i)\geq 0$ for all $i\in[K]$, and $\sum_i\pi(i)=~1$;
		\item[ii.] The vertex set $V=V(G)$ is the disjoint union of $K$ blocks
		$V=\mathcal{B}_1 \sqcup \mathcal{B}_2   \sqcup \cdots \sqcup \mathcal{B}_K$, where each vertex $v\in V$ is independently assigned to a block according to a Multinomial($1,\pi$) distribution.
		For each vertex $v\in V(G)$, let $b_v$ be the block that $v$ is assigned to.
		\item[iii.] The block probability matrix $\Lambda\in[0,1]^{K\times K}$ is a symmetric matrix.
		Conditional on the block assignment vector $\vec b=(b_v)$, for each pair
		of vertices $\{u,v\}\in\binom{V}{2}$, 
		$$\left\{\mathds{1}\{u\sim_{G} v\}\right\}\stackrel{\text{ind.}}{\sim}\text{Bernoulli}(\Lambda[b_u,b_v]).$$
	\end{itemize}
\end{definition}
As in \cite{Peixoto_HSBM}, we will define the HSBM recursively from the top down.
In essence, the 2-level HSBM is an SBM where each block itself has an SBM structure.
If further, every one of the blocks of the second level is an SBM then we have a three level HSBM, and so on.
Formally, we have the following recursive definition.
\begin{definition}
	We say that an $n$-vertex random graph $G$ is an instantiation of a 2-level hierarchical stochastic blockmodel with parameters $$(n,K_1,\Lambda_1,\pi_1,\{K_2^{(j)},\Lambda_2^{(j)},\pi_2^{(j)} \}_{j=1}^{K_1})$$ if
	\begin{itemize}
		\item[i.] The block membership vector $\pi_1\in\mathbb{R}^{K_1}$ satisfies $\pi_1(i)\geq 0$ for all $i\in[K_1]$, and $\sum_i\pi_1(i)=~1$;
		\item[ii.] The vertex set $V=V(G)$ is the disjoint union of $K_1$ blocks
		$V=\mathcal{B}^1_1 \sqcup \mathcal{B}^1_2   \sqcup \cdots \sqcup \mathcal{B}^1_{K_1}$, where each vertex $v\in V$ is independently assigned to a block according to a Multinomial($1,\pi_1$) distribution.
		For each vertex $v\in V(G)$, let $b^{(1)}_v$ be the block that $v$ is assigned to.
		\item[iii.] The block probability matrix $\Lambda_1\in[0,1]^{K_1\times K_1}$ is a hollow symmetric matrix.
		Conditional on the block assignment vector $\vec b^{(1)}=(b^{(1)}_v)$, for each pair
		of vertices $\{u,v\}\in\binom{V}{2}$ with $b^{(1)}_u\neq b^{(1)}_v$, 
		$$\{\mathds{1}_{u\sim_{G} v}\}\stackrel{\text{ind.}}{\sim}\text{Bernoulli}(\Lambda_1[b^{(1)}_u,b^{(1)}_v]).$$
		\item[iv.] 
		For each $j\in[K_1]$, conditional on the block assignment vector 
		$\vec b^{(1)}=(b^{(1)}_v)$ 
		and $|\mathcal{B}^1_j|>0$, 
		we have that 
		$$G[\mathcal{B}^1_j]\sim SBM(|\mathcal{B}^1_j|,K_2^{(j)},\Lambda_2^{(j)},\pi_2^{(j)} ).$$
		Moreover, conditional on the block assignment vector 
		$\vec b^{(1)}=(b^{(1)}_v)$
		the collection 
		$\{G[\mathcal{B}^1_j]\}_{j=1}^{K_1}$ are mutually independent.
	\end{itemize}
\end{definition}
\noindent Formally defining the HSBM beyond the second level of the hierarchy is notationally complex, but the main idea is that a $k$-level HSBM has the same set-up as the $2$-level HSBM except that at part \emph{iv.} of the definition, we have
\begin{itemize}
	\item[\emph{iv'.}] 
	For each $j\in[K_1]$, conditional on the block assignment vector 
	$\vec b^{(1)}=(b^{(1)}_v)$ 
	and $|\mathcal{B}^1_j|>0$, 
	we have that 
	$$G[\mathcal{B}^1_j]\sim \text{k-1 level HSBM}.$$
	Moreover, conditional on the block assignment vector 
	$\vec b^{(1)}=(b^{(1)}_v)$, we have that 
	the collection 
	$\{G[\mathcal{B}^1_j]\}_{j=1}^{K_1}$ are mutually independent.
\end{itemize}
Recursively applying \emph{iv'.} together with the definition of a 2-level HSBM allows us to define HSBMs of arbitrary depth ($\leq n$ of course).

\begin{remark}
	\emph{ For a $k$-level HSBM, we shall adopt the notation from the definition of hierarchical graphs already established.  Namely, we will denote the $j$-th block at level $i$ via $\mathcal{B}_j^i$, where the blocks are labeled in order beginning with those of $B^{i-1}_{(1)}$ and ending with those of $B^{i-1}_{(n_{i-1})}$.  We will denote the block-membership function at level $i$ via $\vec b^{(i)}$.
	}
\end{remark}

While the model complexity of the HSBM can grow quite rapidly as we allow for more structure in each block of the higher level SBMs, the top-down structure does offer complexity savings versus an SBM with the same number of blocks at the bottom-level of the hierarchy. 
For example, the 1-level HSBM (i.e., the SBM) with $K$ blocks requires 
$$
\underbrace{\binom{K}{2}+K}_{\text{for }B}+\underbrace{K-1}_{\text{for }\pi}=O(K^2)
$$
parameters to define, 
while the $2$-level HSBM requires
\begin{align*}
\underbrace{\binom{K_1}{2}}_{\text{for }B_1}\!+\!\underbrace{K_1-1}_{\text{for }\pi_1}&+\!\!\sum_{j=1}^{K_1}\Bigg[\underbrace{\binom{K_2^{(j)}}{2}\!+\!K_2^{(j)}}_{\text{for }B_2^{(j)}}+\underbrace{K_2^{(j)}-1}_{\text{for }\pi_2^{(j)}}\Bigg]\\
&=O(K_1^2+\sum_j (K_2^{(j)})^2)
\end{align*}
parameters.
As a simple example, if $K_1=3$ and each $K_2^{(j)}=3$, then the HSBM has 9 blocks at its bottom level and requires 29 parameters while a 9  block SBM requires 53 parameters.
The complexity of an HSBM can be further reduced by enforcing repeated motif structure \cite{lyzinski2015community}, though we do not explore this further herein.
\begin{remark}
	\emph{Consider a $k$-level HSBM $G$.  
		Letting $(n_i)_{i=1}^k$ denote the number of blocks at level $i$ in the hierarchy of $G$, so that (for example)
		\begin{align*}
			n_1&=K_1\\
			n_2&=\sum_{i=1}^{K_1} K_2^{i}\\
			n_3&=\sum_{i=1}^{K_1}\sum_{j=1}^{K_2^{i}}K_3^{(j,i)}\\
			n_4&=\sum_{i=1}^{K_1}\sum_{j=1}^{K_2^{i}}\sum_{\ell=1}^{K_3^{(j,i)}}K_4^{(\ell,j,i)}\\
			&\vdots
		\end{align*}
		(where $K_3^{(j,i)}$ is the number of blocks in the $j$-th block of the $i$-th block of $G$; $K_4^{(\ell,j,i)}$ is the number of blocks in the $\ell$-th block of the $j$-th block of the $i$-th block of $G$; etc...) and conditioning on all blocks being non-empty at each level of the hierarchy, $G$ defines a distribution over the subset of $k$-level hierarchical graphs
		$$\hg_n^{\vec n}:=\{(g,H)\in\hg_n\,|\text{ the signature of \emph{H} is }\vec n\}.$$
		We then have that (for example), $H(v,i)=\vec b^{(i)}_v$ for all $v\in [n]$ and $i\in[k]$.}
\end{remark}

\subsubsection{Hierarchical Subgraph Nomination}

A key component to the hierarchical subgraph nomination (HSN) inference task is the notion of subgraphs of interest within and across the network pair.
This is easily accomplished in non-random settings, where a fixed set of indices can be used to define the subgraphs of interest within and across networks.  
In order to expedite this in random networks, for a fixed pair of signatures $\vec n$ and $\vec m$, we will consider distributions restricted to $\hgn\times\hgm$; this will allow us to define a consistent set of indices for subgraphs of interest across the random network pair.

In hierarchical subgraph nomination, we consider a pair of hierarchical graphs $(g_1,H_1)\in\hgn$ and $(g_2,H_2)\in\hgm$ where $H_1$ and $H_2$ are unobserved and only the graphs $g_1$ and $g_2$ are observed.
We are additionally given a set of ``training'' subgraphs of interest in $g_1$, denoted
$$ T_1=\{B^{s_{1}}_{s_{2},1} \}_{s=(s_1,s_2)\in S_1}$$
where $S_1$ is a set of ordered pair indices $S_1=\{s=(s_{1},s_{2})\}$ of the given subgraphs of interest,
with $s_{1}$ denoting the level and $s_{2}$ the index of the subgraph in $H_1$. 
Note that the second index in the subscript of  $B^{\bullet}_{\bullet,1}$ is used to denote the subgraph in $g_1$, whereas $B^{\bullet}_{\bullet,2}$ will be used to indicate subgraphs in the hierarchy of $g_2$.
Note that the possible indices of the seed graphs depends on the structure of $H_1$, and so we will explicitly tether these two in the sequel, writing
$((g_1,H_1),S_1)$, and writing $\hgs_n^{\vec n}$ for the collection of all such feasible triples.

The aim then is to 
\begin{itemize}
	\item[1.] First estimate the latent hierarchical structure $H_2$ of $g_2$; we will denote this estimate via $\widehat H_2$, and we will let 
	$$\widehat B_{j,2}^i=\{v\in V(g_2)|\widehat H_2(v,i)=j\}.$$
	Let the collection of all subgraphs in the partition defined by $\widehat H_2$ be denoted $\widehat B$.
	\item[2.] Compute dissimilarity measurements
	$$\bD:T_1\times \widehat B\mapsto[0,1]$$
	between each subgraph of interest in $(g_1,H_1)$ and each subgraph defined by $\widehat H_2$.
	Note that $\bD$ ideally is a function that can compute the dissimilarity between any graph of order up to $n$ and any graph of order up to $m$; and the larger the value of $\bD$ between two graphs, the more dissimilar the networks.
	As $\bD$ needs to compute dissimilarities across graphs of different sizes and orders, it may encompass a collection of dissimilarity measures. 
\end{itemize}
After a few preliminary definitions, we will be ready to define a HSN scheme.
\begin{definition}
	For a $k$-level hierarchical graph $(g,H)$ and $i\in[k]$, let $\calT_{g,i}^H$ denote the set of all total orderings of $B^i$.  Let $\calT_{g}^H=\bigotimes_{i=1}^{k} \calT_{g,i}^H$.
\end{definition}
\noindent For example, in the hierarchical subgraph depicted in Figure \ref{fig:hier_ex}, we have
\begin{align*}
\calT_{g,2}^H=\left\lbrace
\begin{bmatrix}
	B_{1}^2\\
	B_{2}^2\\
	B_{3}^2
\end{bmatrix}\right.,
\begin{bmatrix}
	B_{1}^2\\
	B_{3}^2\\
	B_{2}^2
\end{bmatrix},
&\begin{bmatrix}
	B_{2}^2\\
	B_{1}^2\\
	B_{3}^2
\end{bmatrix},\\
\begin{bmatrix}
	B_{2}^2\\
	B_{3}^2\\
	B_{1}^2
\end{bmatrix},
\begin{bmatrix}
	B_{3}^2\\
	B_{1}^2\\
	B_{2}^2
\end{bmatrix},
&\left.\begin{bmatrix}
	B_{3}^2\\
	B_{2}^2\\
	B_{1}^2
\end{bmatrix}
\right\}    
\end{align*}

where
$$
\begin{bmatrix}
	B_{i}^2\\
	B_{j}^2\\
	B_{k}^2
\end{bmatrix}$$
is shorthand for the ordering 
$$
B_{i}^2>
B_{j}^2>
B_{k}^2.
$$
An example element of $\calT_{g}^H$ is given by
\begin{align*}
    \left([B^1_1],\,\right. &[B_{3}^2,B_{2}^2,B_{1}^2],\\ &\left.[B^3_2,B^3_9,B^3_8,B^3_1,B^3_3,B^3_6,B^3_5,B^3_7,B^3_4]\right).
\end{align*}
As in the case of vertex nomination, if vertex labels in $(g_2,H_2)$ are uninformative, then an HSN must account for the indistinguishability of subgraphs with isomorphic structure.
To wit, we have the following definition: 
For a $k$-level hierarchical network $(g,H)$ and $i\in[k]$, $j\in[n_i]$, we define
\begin{align*}
	\mathfrak{I}(B_j^i;g)&=\\
	\{&\ell\in[n_i]\big|\,
	B_\ell^i\subset B_{(j)}^{i-1}\text{ and there exists an }\\
	&\text{ automorphism }\sigma\text{ of }g\text{ with }\sigma(B^i_\ell)=B_j^i \}.
\end{align*}
\noindent These are indices of the elements of the partition at level $i$ that are indistinguishable from  $B_j^i$ without further information/supervision, and it is sensible to require that a HSN rank all these subgraphs as equally interesting.
In vertex nomination, accounting for label uncertainty was achieved by means of an obfuscation function; in HSN, the uncertainty is not at the level of labels so much as at the level of subgraphs and this 
this can be achieved by further requiring the dissimilarity $\bD$ satisfies the following condition: 
\begin{align} 
         \bD&(B^{s_1}_{s_2,1},\cdot)\text{ is constant over graphs indexed by}\notag\\ &\mathfrak{I}(\widehat B_{j,2}^i;g_2)\text{ for each }s=(s_1,s_2)\in S_1.\label{eq:D}
    \end{align}

We are now ready to define hierarchical subgraph nomination schemes formally.
\begin{definition}{\emph[Hierarchical Subgraph Nomination Scheme (HSN)]}\label{def:SGN}
	Let $((g_1,H_1),S_1)\in\hgsn$ and $((g_2,H_2),S_2)\in\hgsm$ and let $T_1$ be the parts of $H_1$ indexed by $S_1$.
	A \emph{Hierarchical subgraph nomination scheme} is composed of two parts:
	\begin{itemize}
		\item[i.] An estimator $\widehat H_2=\widehat H_2(g_1,g_2,T_1)$ of the hierarchical clustering of $g_2$ provided by $H_2$; let the signature of $\widehat H_2$ be denoted $\hat m_2=(\hat m_{k,2})$;
		\item[ii.] A dissimilarity $
		\bD$ satisfying Eq.\@ (\ref{eq:D}) which produces a ranking scheme $\Phi_{n,m}$, where
		$$\Phi_{n,m}\left(g_1,g_2,T_1,\widehat{H}_2 \right)\in\calT_{g_2}^{\widehat H_2}.
		$$ 
		For each level $i$ of $\widehat H_2$, the subgraphs are ordered in $\Phi_{n,m}$ via increasing value of 
		$$\min_{s=(s_1,s_2)\in S_1} \bD(B^{s_1}_{s_2,1},\widehat B_{j,2}^i),$$ 
		with ties broken in a fixed but arbitrarily manner.
	\end{itemize}
\end{definition}


\subsection{Loss in HSN Schemes}
\label{HSGNBEBO}

The goal of HSN is to effectively query $g_2$ given limited training resources from $g_1$.
Given the resources required for a user to verify the interestingness of the returned subgraphs,
we seek to maximize the probability that a priori unknown subgraphs of interest in $g_2$ are close to the top of the returned rank list. 

For evaluation purposes, it is necessary (at least in theory) to be able to compare the true-but-unknown subgraphs of interest in $((g_2,H_2),S_2)\in\hgs_m$ with elements of the HSN ranked list.
In order to account for the fact that the dissimilarity used to construct the HSN scheme may be mis-specified (i.e., not captured exactly), the dissimilarity for verification will be denoted $\bD_E$ (the dissimilarity used for evaluation that defines the true but unknown subgraphs of interest), though we do not discount the possibility that $\bD_E=\bD$ for any given scheme.

A logical loss function for HSN is motivated by the concept of precision  in information retrieval.
As the goal of HSN is to efficiently query large networks for structures of interest, at a given level $k$, considering precision at $i$ for $1<i\ll \hat m_{k,2}$ 
enables us to model the practical loss associated with using a HSN scheme to search for $\{B^{s_1}_{s_2,2}\}_{(s_1,s_2)\in S_2}$ in $(g_2,H_2)$ given limited resources.
There are two levels to the loss function for a given scheme $\Phi_{n,m} =(\widehat H_2,\bD)$:
\begin{itemize}
	\item[i.]  The error in approximating $H_2$ via $\widehat H_2$;
	\item[ii.] Given $\widehat H_2$, the potential mismatch between $\bD$ and $\bD_E$.
\end{itemize}
Our loss function will account for these error sources as follows.
Let $F$ be a distribution supported on $\hgs_n\times \hgs_m$, and let 
$$
((G_1,H_1),(G_2,H_2))\sim F.
$$
Consider fixed subgraph of interest index sets $S_1$ for $G_1$ and $S_2$ for $G_2$, and let $T_1$ (resp., $T_2$) be those subgraphs in $(G_1,H_1)$ (resp., $(G_2,H_2)$) indexed by $S_1$ (resp., $S_2$).
Let $\Phi_{n,m}=(\widehat H_2,\bD)$ be an HSN scheme and let the ranking provided at level $k$ of $\Phi_{n,m}$ be denoted via $\Phi_{n,m}^k$ (with the implicit assumption that this is an empty list if $\hat m_{k,2}=0$).
\begin{definition}[HSN loss function, level-$(i,k)$ error at hierarchical level]
	\label{def:lossfcn}
	With setup as above, for $i,k<n$ we define the \emph{level-$(i,k)$ nomination loss with threshold $t>0$ under dissimilarity $\bD_E$} via (where to ease notation, we will use 
	$\Phi_{n,m}^k$ for $\Phi_{n,m}^k(g_1,g_2,T_1,\widehat H_2)$ and $\Phi_{n,m}^k[j]$ the $j$-th ranked subgraph in this list)
	$$\ell_{i,k,t}:=\ell_{i,k,t}(\Phi_{n,m},g_1,H_1,S_1,g_2,H_2,S_2,\bD_E)
	$$
	where if we let $J_{i,k} =\min(i,\hat{m}_{k,2})$ and $\mathfrak{W}_{\Phi_{n,m}[j]}=\left\{
			\bigcap_{(s_1,s_2)\in S_2}
			\{\Delta_{E}(\Phi_{n,m}[j],B^{s_1}_{s_2,2})>t\}\right\}$
	\begin{align} 
		\label{eq:lossfcn}
		\ell_{i,k,t}:=\begin{cases}\frac{1}{J_{i,k}}
			\sum\limits_{j=1}^{J_{i,k}}\mathds{1}\left\{\mathfrak{W}_{\Phi_{n,m}[j]}\right\}&\text{ if }\hat m_{k,2}>0\\
			1&\text{ if }\hat m_{k,2}=0
		\end{cases}
	\end{align}
	For distribution $F$ as above, the \emph{level-$(i,k)$ error with threshold $t>0$ under dissimilarity $\bD_E$} of $\Phi_{n,m}$ for recovering $S_2$
	is defined to be $L_{i,k,t}=\mathbb{E}_{F}(\ell_{i,k,t})$.  
	Letting $\mathfrak{H}_{n,m}$ be the collection of all HSN schemes (i.e., the set of all ordered pairs of estimators and dissimilarities $(\widehat H_2,\bD)$), the Bayes optimal level-$(i,k)$ scheme with threshold $t>0$ under dissimilarity $\bD_E$ is defined to be any scheme that achieves the Bayes' error, which here is defined to be 
	$$\mathrm{min}_{\Phi_{n,m}\in \mathfrak{H}_{n,m}} \mathbb{E}_{F}(\ell_{i,k,t}(\Phi_{n,m}))$$
\end{definition}
\noindent

\noindent While the number of possible dissimilarities is indeed uncountably infinite, there are nonetheless a finite number of possible rankings that can be achieved via a combination of $\widehat H_2$ and $\bD$ (hence $\min$ rather than $\text{inf}$ in the Bayes error definition), and so the Bayes error is indeed achieved by at least one such pair.


\subsection{User-in-the-loop Supervision}
\label{sec:user}

Interactive machine learning, via incorporating user-in-the-loop supervision, can lead to an enhanced user experience and better downstream inference performance \cite{amershi2014power}.
In subgraph nomination, the need for user-in-the-loop supervision can be understood as follows.
Consider nominating in $((g_2,H_2),S_2)$ at level $k$, where $S_2=\{(k,\ell)\}$, and  
$|\mathfrak{I}(B^{k}_{\ell,2};g_2)|>1$.
Even if $\widehat H_2$ agrees with $H_2$ at level $k$, there are multiple subgraphs in $(g_2,H_2)$ that are isomorphic to the unknown subgraph of interest, and the HSN scheme has no information available to distinguish these.
In this setting, it is reasonable to model the relative order of the elements in $\mathfrak{I}(B^{k}_{\ell,2};g_2)$ in an HSN scheme as (effectively) uniformly random.
For example, if $|\mathfrak{I}(B^{k}_{\ell,2};g_2)|=c$, then considering top $h$ of our nomination list ($h<c$), a scheme that identifies the correct structure for $B^{k}_{\ell,2}$ would still have
probability 
$$
\frac{\binom{c-1}{h}}{\binom{c}{h} }=\frac{c-h}{c}=1-\frac{h}{c}
$$
of not finding $B^{k}_{\ell,2}$.
This can be mitigated by the following user-in-the-loop system:
\begin{itemize}
	\item[i.] The user can take a limited number of single vertex inputs and can output whether that vertex is interesting or not (i.e., part of an interesting subgraph).  This output can be modeled as error-free (oracle-user) or errorful.
	\item[ii.] The supervision can then be used to re-rank the subgraphs in the nomination scheme.
\end{itemize}
While practically, we do not suspect that there will be many perfect matches to the subgraphs of interest in the hierarchy provided by $H_2$, it may be the case that there are many subgraphs ``close" to the subgraph of interest in the (possibly lossy) estimate $\widehat H_2$, in which case this supervision is equally necessary.

What follows is a formalization of the above heuristic.
We first define the concept of a user-in-the-loop; noting that our definition is not the most general, as we focus in our analysis on binary users; i.e., they can only output $1$ (yes) or $0$ (no) for the interestingness of a vertex. 
In general, one might allow for the user to use a rating system with more levels, or even add a continuous space for responses. 
\begin{definition}[VN User]
	Let $(g_1,H_1)\in\hgs_n$ with indices of interest $S_1$ and $(g_2,H_2)\in\hgs_m$ with indices of interest $S_2$.
	Let $(\theta,\gamma)\in [0,1]\times[0,1]$ and $t\in\mathbb{Z}>0$ with $t<m$.
	We define the capacity $t$, \emph{HSGN user-in-the-loop} for $((g_2,H_2),S_2)$ with parameters $(\theta,\gamma)$ ( denoted $U_t$) as follows:
				\begin{figure*}[t!]
	\centering
	\includegraphics[width=0.8\textwidth]{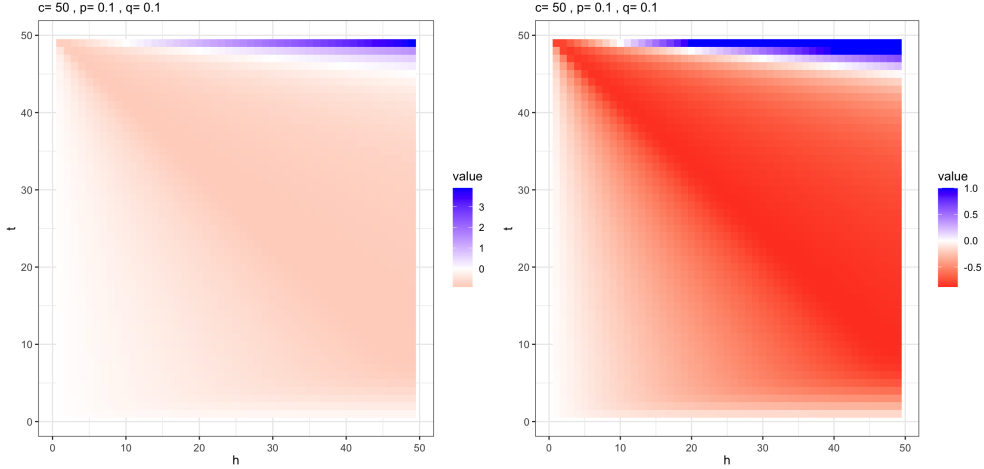}
	\caption{For $c=50$ we plot (for $h,t\in\{1,2,\ldots,49\}$) the relative loss $R(c,t,h,p,q)$ in the setting of Theorem \ref{thm:losspq} part (iii) in the left panel $\min(R(c,t,h,p,q),1)$ in the right panel.}
	\label{fig:numericL}
\end{figure*}
	\begin{itemize}
		\item[i.] Letting $\Omega$ be the underlying sample space, the user is a function $$U_t:\mathcal{T}_{V_2,t}\times\Omega\mapsto\{0,1\}^t$$ 
		where $\mathcal{T}_{V_2,t}$ is the set of ordered $t$-tuples of distinct elements from $V_2=V(g_2)$.
		Note that dependence on $\Omega$ will be suppressed when appropriate.
		\item[ii.]
		For each $\eta\in\mathcal{T}_{V_2,t}$ and $x\in\{0,1\}^t$, let $\mathfrak{F}_{\in,i}=\{\eta_i\in \bigcup_{(s_1,s_2)\in S_2}B_{s_2,2}^{s_1}\} $, $\mathfrak{F}_{\notin,i}=\{\eta_i\notin \bigcup_{(s_1,s_2)\in S_2}B_{s_2,2}^{s_1}\} $, define
		\begin{align}\label{eq:user}
			\mathbb{P}(U_t(\eta)&=x)=\notag\\
			\prod_{i=1}^t\bigg[\bigg(&\mathds{1}\{\mathfrak{F}_{\in,i}\} \theta^{x_i}(1-\theta)^{1-x_i}  \bigg)\notag\\
			+&\bigg(\mathds{1}\{\mathfrak{F}_{\notin,i}\} \gamma^{x_i}(1-\gamma)^{1-x_i}\bigg)\bigg]
		\end{align}
		In essence, the user has independent binary components where the Bernoulli success probabilities depend only on the membership (or lack thereof) in a subgraph of interest.
	\end{itemize}
	For a given $((g_2,H_2),S_2)\text{ and }t$, we define the set of all such users by $\mathfrak{U}_t=\mathfrak{U}_{((g_2,H_2),S_2),t}$.
\end{definition}
Effectively, the HSN user-in-the-loop outputs a sequence of $\{0,1\}$ values, one for each vertex in a training set $\eta$. 
An output of 1 is considered the positive answer, i.e. an interesting vertex; for vertices in $\cup_{(s_1,s_2)\in S_2}B_{s_2,2}^{s_1}$, this is a correct answer, and an error otherwise.
An output of 0 is the negative answer meaning that this vertex is not of interest; 
for vertices not in $\cup_{(s_1,s_2)\in S_2}B_{s_2,2}^{s_1}$, this is a correct answer and is an error otherwise.
This allows us to simultaneously model oracle users ($\theta=1$, and $\gamma=0$) and errorful users ($\theta<1$, and $\gamma>0$).
The capacity of the user (i.e., $t$) allows us to model the practical setting in which user-in-the-loop resources are costly and only limited supervision is available.
\begin{figure*}[t!]
	\centering
	\includegraphics[width=1\textwidth]{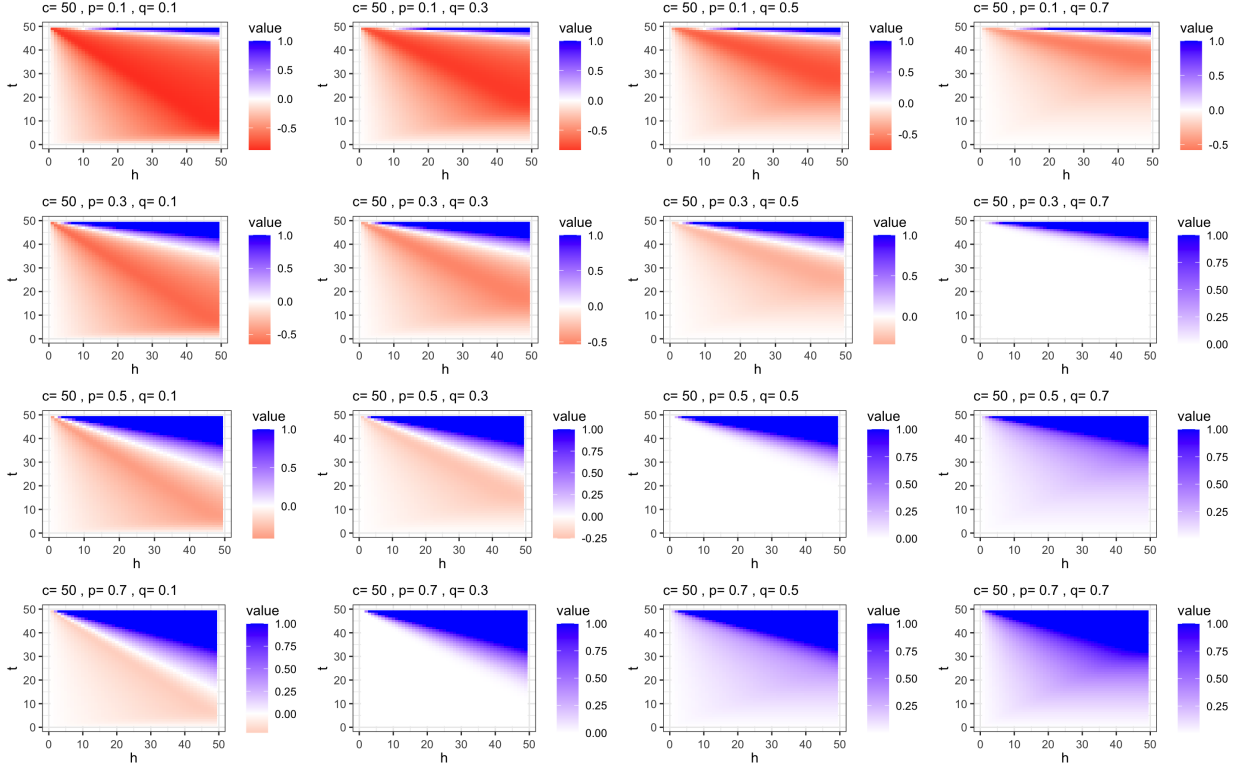}
	\caption{For $c=50$ we plot (for $h,t\in\{1,2,\ldots,49\}$) $\min(R(c,t,h,p,q),1)$ for the loss function in the setting of Theorem \ref{thm:losspq} part (iii) and $R$ defined in Equation (\ref{eq:R}).
		Different panels correspond to different values of $p$ and $q$.}
	\label{fig:numericL2}
\end{figure*}
Users can be incorporated into HSN schemes as follows.  
\begin{definition}{User Aided HSN Scheme (UHSN)}
	Consider the setting in Definition (\ref{def:SGN}). 
	Let $t\leq n_j^{(2)}$ and $U_t\in\mathfrak{U}_t$.
	Consider an HSN $\Phi_{n,m}=(\widehat H_2,\bD)$, and let $\eta$ be an ordered $t$-tuple of distinct elements of $V_2=V(g_2)$.
	For each $k$, 
	$U_t$ acts on $\Phi_{n,m}^k$ as follows:
	\begin{itemize}
		\item[i.] Given $\eta$, $U_t$ is distributed according to Eq. \ref{eq:user}.  Let the output of the user process be denoted $x\in\{0,1\}^t$.
		\item[ii.] Let
		\begin{align*}
			I_k:=&\{\ell\in [\hat m_k]\text{ s.t. more than half}\\
			&\text{ of the vertices in }\eta\cap\widehat B_{\ell,2}^{k}\\
			&\text{ were labeled interesting (i.e., as 1)} \\&\text{ by the user}\}\\
			N_k:=&\{\ell\in [\hat m_k]\text{ s.t. at least half }\\
			&\text{ of the vertices in }\eta\cap\widehat B_{\ell,2}^{k}\\
			&\text{ were labeled as not interesting (i.e., as 0)}\\&\text{ by the user}\}
		\end{align*}
		If $\eta\cap\widehat B_{\ell,2}^k=\emptyset$, then $\ell\notin I_t\cup N_t$.
		\item[iii.] 
		Let the indices of $I_t$ be indexed via $(\mathfrak{i}_1,\cdots,\mathfrak{i}_{|I_t|})$ where (according to $\Phi_{n,m}^k$)
		$$
		\widehat B_{\mathfrak{i}_1,2}^{k}>
		\widehat B_{\mathfrak{i}_2,2}^{k}>\cdots>
		\widehat B_{\mathfrak{i}_{|I_t|},2}^{k}.
		$$
		Similarly index $N_t$ via $(\mathfrak{n}_1,\cdots,\mathfrak{n}_{|N_t|})$, and
		$M_t:=[\hat m_k]\setminus \{I_t,N_t\}$ (the unsupervised subgraph indices) via 
		$(\mathfrak{m}_1,\cdots,\mathfrak{m}_{|M_t|})$
		The user improved ranking is then given by
		\begin{align*}
			\Phi_{n,m}^{k,U}=
			\bigg(
			\widehat B_{\mathfrak{i}_1,2}^{k}&>
			\widehat B_{\mathfrak{i}_2,2}^{k}>
			\widehat B_{\mathfrak{i}_{|I_t|},2}^{k}>
			\widehat B_{\mathfrak{m}_1,2}^{k}>\cdots\\
			\cdots>&\widehat B_{\mathfrak{m}_2,2}^{k}>
			\widehat B_{\mathfrak{m}_{|M_t|},2}^{k}>
			\widehat B_{\mathfrak{n}_1,2}^{k}>\cdots\\
			\cdots>&
			\widehat B_{\mathfrak{n}_2,2}^{k}>\widehat B_{\mathfrak{n}_{|N_t|},2}^{k}
			\bigg).
		\end{align*}
	\end{itemize}
\end{definition}
\begin{remark}
	\emph{
		We note here the possibility of even an oracle user introducing large errors into a ranking scheme based on an approximate $\widehat H_2$.  Indeed, even if $\widehat B_{\ell,2}^k$ is very similar to an interesting subgraph in $g_1$ according to $\mathbf{\Delta}$, it is still possible that uninteresting vertices are chosen to provide to the user and $\ell\in N_t$.  This can be mitigated by choosing multiple vertices from each of some of the top ranked subgraphs for the user to evaluate.}
\end{remark}

\begin{remark}
	\emph{
		For iterative applications of the user-in-the-loop, we can sequentially run the user-in-the-loop, first on the output of $\Phi_{n,m}$, then on $\Phi_{n,m}^{U}$ (with $t$ new user training points), and so on.}
\end{remark}


\subsubsection{The (Theoretical) Benefit of the User-in-the-loop}
\label{sec:use-theory}
\paragraph{}
We turn our attention now to take a look at the theoretical benefit of the user-in-the-loop in the setting of HSN.
As in the motivating example for the user, the setting for this section will be as follows.
Letting 
$(g_1,H_1)\in\hgs_n$ and $(g_2,H_2)\in\hgs_m$ with respective interesting index sets $S_1$ and $S_2$, consider a HSN scheme $\Phi_{n,m}$ with $\widehat H_2=H_2$.
Consider the nomination provided by $\Phi^k_{n,m}$ and the simple setting in which $S_2=\{(k,\ell)\}$ with
$|\mathfrak{I}(B_{\ell,2}^k;g_2)|=c>1$.
In this setting, it is reasonable to model the relative order of the elements in $\mathfrak{I}(B_{\ell,2}^k;g_2)$ in an $\Phi^k_{n,m}$ as (effectively) uniformly random (though in practice, they are a fixed arbitrary order).
\begin{figure*}[t!]
	\centering
	\includegraphics[trim={0 3mm 3mm 3mm},clip, width=0.7\textwidth]{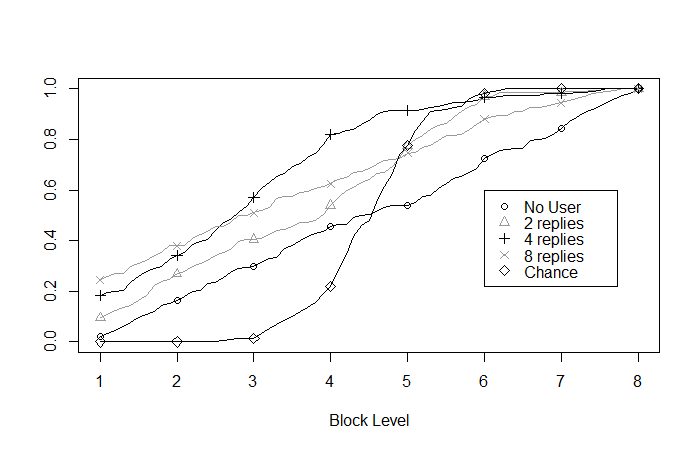}
	\caption{We consider 2000 Monte Carlo replicates of distributions as in Section \ref{sec:model}, and from each distribution, we generate 100 networks.
		We use Method 1, and performance with an oracle user (varying the replies for the user; i.e., the amount of supervision provided) and the chance algorithm are compared. 
		In all cases, we plot the proportion of trials (y-axis) that ranked the correct block in the nth place (x-axis) which represents $1-L_n(\phi,B^*)$.}
	\label{fig:sim1-1}
\end{figure*}

\begin{theorem}
	\label{thm:losspq}
	Given the setup above where the scheme $\Phi_{n,m}$ effectively ranks the elements of $\mathfrak{I}(B_{\ell,2}^k;g_2)$ uniformly at random at the top of the rank list, let $t\leq c<m_k$ be the capacity of the user-in-the-loop.
	Consider any training set for the user where $\eta$ contains exactly one element of each $\Phi_{n,m}^k[h]$ with $1\leq h\leq t$.
	Let $E_h$ denote the event that $\Phi_{n,m}^U$ ranks $B_{\ell,2}^k$ not in the top $h$.
	Considering $c>t,h$, we have the following.
	\begin{itemize}
		\item[i.] 
		Let $U_t$ be an oracle user, then $\mathbb{P}(E_h)=\max(1-\frac{h+t}{c},0).$
		\item[ii.] 
		Let $(\theta,\gamma)=(1-p,0)$ for  $p<1$, then 
		\begin{itemize}
			\item If $t+h\leq c$, then
			$\mathbb{P}(E_h)=1-\frac{h}{c}-(1-p)\frac{t}{c}$; note that this is strictly less than $1-h/c$ for all $p\in(0,1)$.
			\item If $t+h> c$, then
			$\mathbb{P}(E_h)=\frac{tp}{c}$; note that if $p<(c-h)/t$, this is strictly less than $1-h/c$.
		\end{itemize}
		
		\item[iii.] Let $(\theta,\gamma)=(1-p,q)$ for $p<1$ and $q>0$.
		Then, letting $F(i; n,p)$ be the CDF of a Binomial$(n,p)$ random variable evaluated at $i$, we have
		\begin{itemize}
			\item $\text{ If }h+t\leq c,\text{ and }\,t\leq h$, then $\mathbb{P}(E_h)=1-\frac{h}{c}-\frac{(1-p-q)t}{c}$; note that if $p+q<1$, this is less than $1-\frac{h}{c}$.
			\item  $\text{ If }h+t> c,\text{ and }\,t\leq h$, then $\mathbb{P}(E_h)=\frac{pt}{c}+ \frac{1}{c}\sum_{i=1}^{c-h}F(i-1;t,1-q)$; this is upper bounded by $(p+q)t/c$ and if $(p+q)<(c-h)/t$, this is strictly less than $1-h/c$.
			\item $\text{ If }h+t\leq c,\text{ and }\,t> h$, then
			\begin{align*}
				\mathbb{P}(E_h)=&1-\frac{h}{c}-\frac{(1-p)t}{c}\\
				&+\frac{1-p}{c}\sum_{i=1}^{t-h}F(i-1;h+i-1,1-q)\\
				&+\frac{1}{c}\sum_{i=t-h+1}^tF(i-1;t,1-q)
			\end{align*}
			Note that a rough upper bound of this is given by $1-h/c+qt/c-(1-p)h/c$, and if $qt<(1-p)h$, this is strictly less than $1-h/c$.
			\item $\text{ If }h+t> c,\text{ and }\,t> h$, then
			\begin{align*}
				\mathbb{P}(E_h)=&\frac{pt}{c}\\
				&+\frac{1-p}{c}\sum_{i=1}^{t-h}F(i-1;h+i-1,1-q)\\
				&+\frac{1}{c}\sum_{i=t-h+1}^{c-h} F(i-1;t,1-q)
			\end{align*}
			Note that a rough upper bound of this is given by $\frac{(1+q)t-(1-p)h}{c}$, and if $(1+q)t+ph<c$, this is strictly less than $1-h/c$.
		\end{itemize}
	\end{itemize}
\end{theorem}
\begin{figure*}[ht!]
	\centering
	\begin{tabular}{cc}
	\includegraphics[width=7cm]{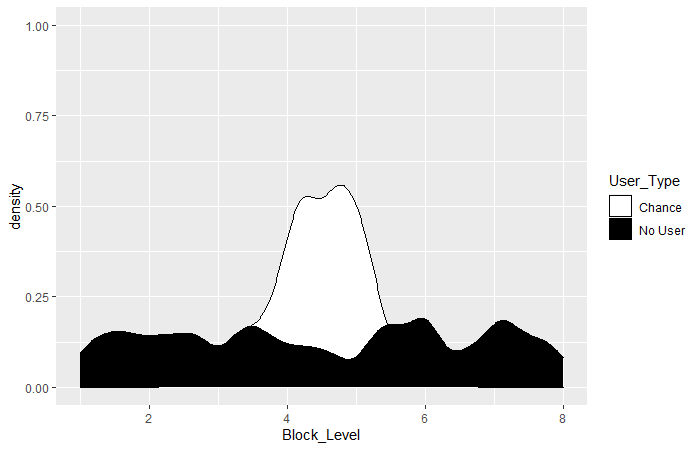}&
	\includegraphics[width=7cm]{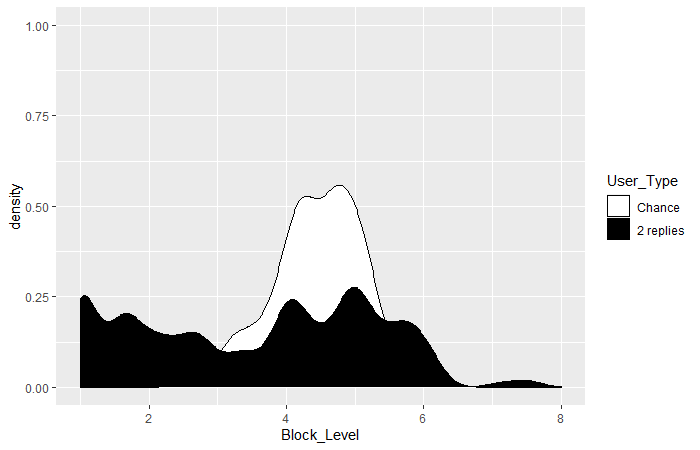}\\
	\includegraphics[width=7cm]{fig5-2.png}&
	\includegraphics[width=7cm]{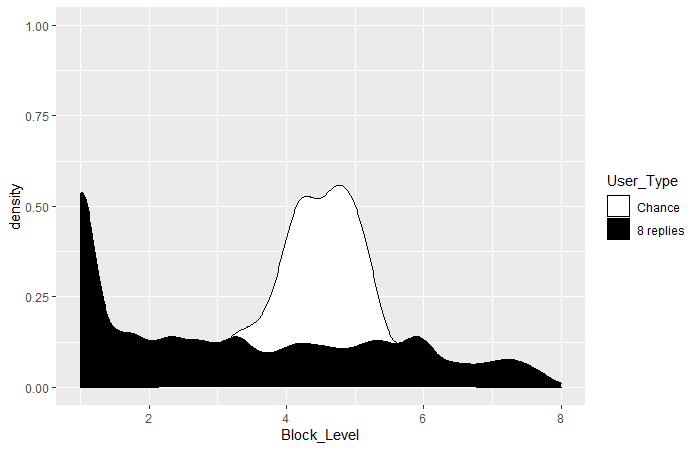}
	\end{tabular}
	\caption{We consider 2000 Monte Carlo replicates of distributions as in Section \ref{sec:model}, and from each distribution, we generate 100 networks.
		We use Method 1, and performance with an oracle user (varying the replies for the user; i.e., the amount of supervision provided) and the chance algorithm are compared. 
		We plot the density of all trials (y-axis) that ranked the correct block in the nth place (x-axis).}
	\label{fig:sim1-2}
\end{figure*}
\noindent We, perhaps surprisingly, see that there are $p$ and $q$ values that guarantee improvement (over the user-in-the-loop-free setting)  in all cases.
The conditions in part $iii.$ bear further consideration.
The interaction of $p,q,h,t$ here is nuanced, and is most easily analyzed numerically, 
as is shown in Figures \ref{fig:numericL} and \ref{fig:numericL2}.
Letting $L(c,t,h,p,q)$ denote the loss from Theorem \ref{thm:losspq} part (iii) (where we can interpret this in the context of Definition \ref{def:lossfcn} by considering $\bD$ as a $0/1$ oracle dissimilarity function), for $c=50$ we plot the relative loss improvement,
\begin{equation}
	\label{eq:R}
	R(c,t,h,p,q)=\frac{L(c,t,h,p,q)-(1-h/c)}{1-h/c},
\end{equation}
for $h$ between 1 and 49 (on the $y$-axis), and $t$ between 1 and 49 (on the $x$-axis).  Different panels correspond to different values of $p$ (varies by row) and $q$ (varies by column). 
Note that in Figures \ref{fig:numericL}, we plot the relative loss in the left panel $R(c,t,h,p,q)$ and $\min(R(c,t,h,p,q),1)$ in the right panel; this truncating aids in distinguishing the areas of improvement from those where $R(c,t,h,p,q)>0$.

In each plot, darker red areas correspond to parameter settings where the loss is improved with user-supervision, 
and darker blue areas to parameter settings where the loss is increased with user-supervision (due to the user error).
A few trends are clear from the figures.
Unsurprisingly, supervision becomes detrimental (i.e., more supervision leads to more loss) as $p$ and $q$ increase and $q\geq 0.5$, although the loss appears to be more tolerant of higher values of $p$ than it is of $q$. 
In all cases, large values of $t$ and $h$ simultaneously lead to training becoming less effective, though $h<t$ seems to yield resilience to larger user-supervised loss for all $p,q$ pairs.

In the event that the hierarchical clustering is noisily recovered in $g_2$ (i.e., $\widehat H_2\neq H_2$), even the supervision provided by an oracle user-in-the-loop may be worse than no extra supervision.
Indeed, consider the setting where $S_2=\{(k,\ell)\}$ and
$$\alpha=\frac{|\widehat B^k_{\ell,2}\cap  B^k_{\ell,2}| }{| \widehat B^k_{\ell,2}| };\quad 
\beta=\min_{i\neq \ell}\frac{|\widehat B^k_{i,2}\cap  B^k_{\ell,2}| }{| \widehat B^k_{i,2}| },$$
then, adopting the notation and setting above, the probability that $\widehat B^k_{\ell,2}$ is not ranked in the top $h$ after oracle $U_t$ supervision is bounded below by the probabilities in part (iii) of Theorem \ref{thm:losspq} with $p=\alpha$ and $q=\beta$.
In particular, there are values of $\alpha$ and $\beta$ under which this error is worse than $1-h/c$ (the error sans additional supervision).

\begin{figure*}[t!]
	\centering
	\includegraphics[width=0.8\textwidth]{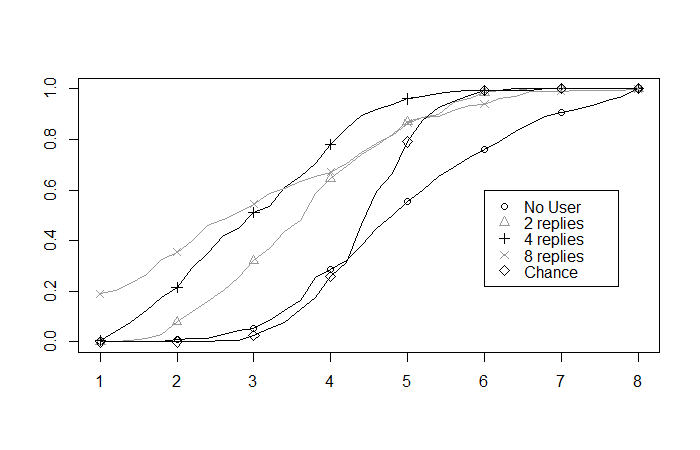}
	\caption{We consider 2000 Monte Carlo replicates of distributions as in Section \ref{sec:model}, and from each distribution, we generate 100 networks.
		We use Method 2, and performance with an oracle user (varying the replies for the user; i.e., the amount of supervision provided) and the chance algorithm are compared. 
		We plot the proportion of trials (y-axis) that ranked the correct block in the nth place (x-axis) which represents $1-L_n(\phi,B^*)$}
	\label{fig:sim2-1}
\end{figure*}

\section{Simulations and Real Data Experiments}

We now provide empirical evidence for the theory we have outlined. 
Namely, we will show through simulations and real data examples the impact of a user-in-the-loop.
We first consider simulations where graphs are drawn from an HSBM distribution. In this setting we evaluate our methodology on the task of nominating subgraphs with a similar motif to our subgraph of interest.
We then consider 57 pairs of analyzed human neural connectomes for which brain regions (i.e., communities) are well defined and the community for each voxel is known. We evaluate our methodology on the task of nominating the same brain region within a given pair of connectomes with varying levels of a user-in-the-loop.

\begin{figure*}[t!]
	\centering
	\begin{tabular}{cc}
	\includegraphics[width=7cm,height=4.5cm]{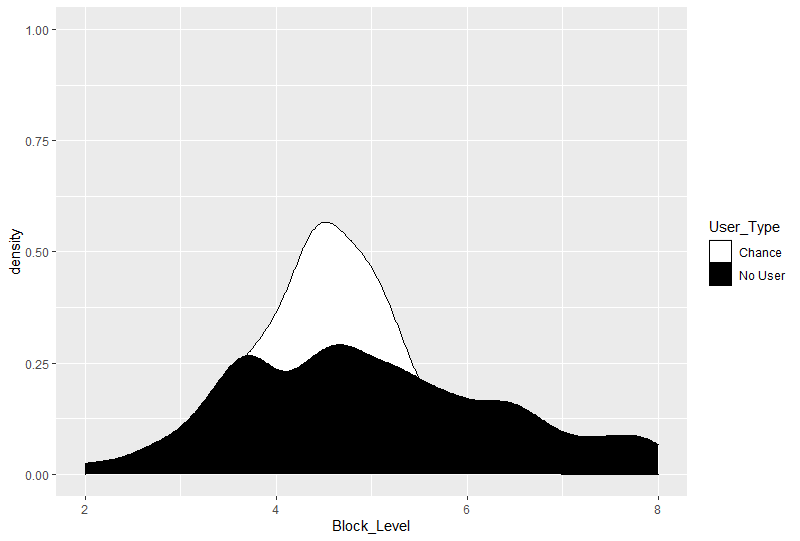}&
	\includegraphics[width=7cm,height=4.5cm]{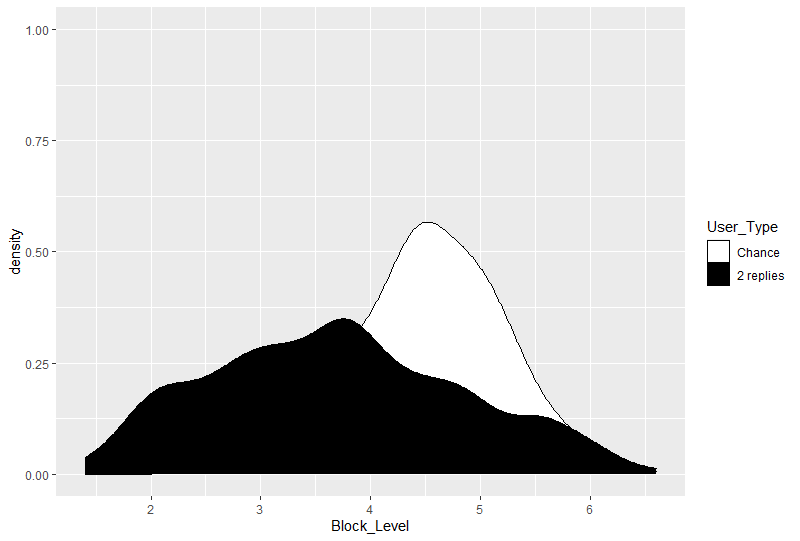}\\
	\includegraphics[width=7cm,height=4.5cm]{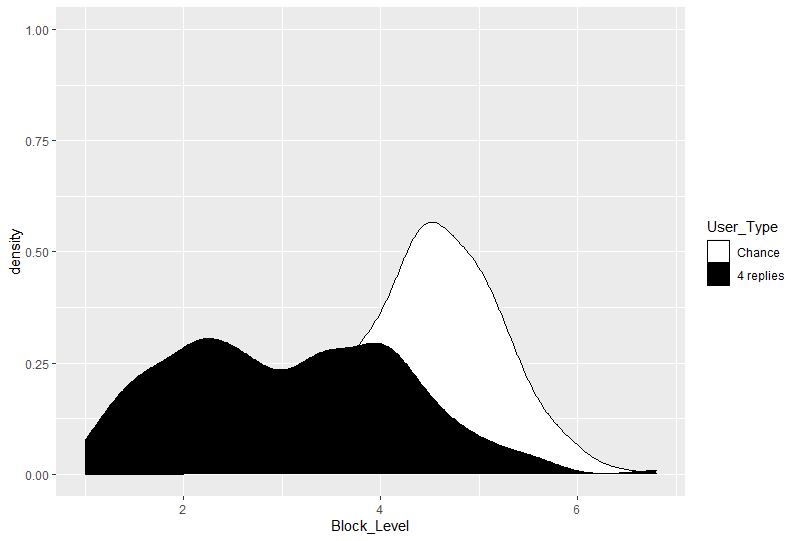}&
	\includegraphics[width=7cm,height=4.5cm]{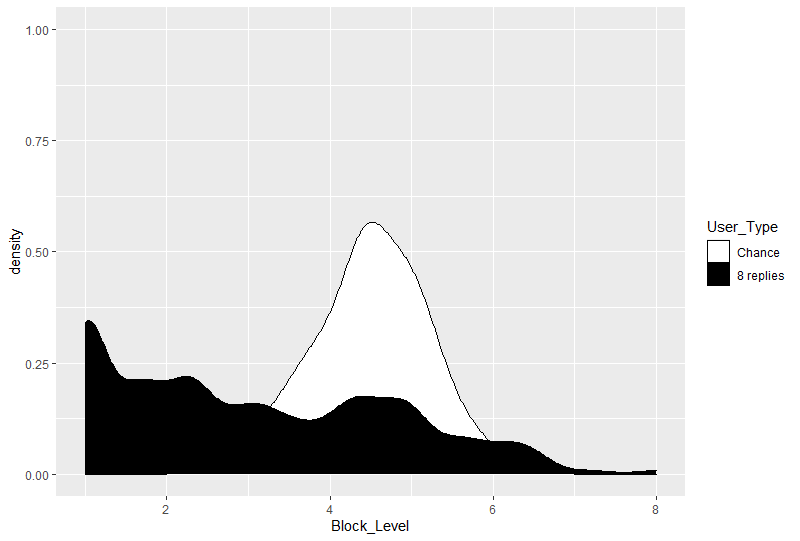}
	\end{tabular}
	\caption{We consider 2000 Monte Carlo replicates of distributions as in Section \ref{sec:model}, and from each distribution, we generate 100 networks.
		We use Method 2 and performance with an oracle user (varying the replies for the user; i.e., the amount of supervision provided) and the chance algorithm are compared. 
		We plot the density of all trials (y-axis) that ranked the correct block in the nth place (x-axis).}
	\label{fig:sim2-2}
\end{figure*}

\subsection{Simulations}
\label{sims}
In this section we will first provide a complete description of the model and procedures that were employed to test our theory in a simulated data setting. 
We then discuss results and how they fit within our theory.

\subsubsection{Model description}
\label{sec:model}
For our first example, we consider nominating within the Hierarchical Stochastic Blockmodel setting of \cite{lyzinski2015community}.
We consider sampling from a 2-level HSBM that has 16 blocks in the first level and 3 sub-blocks in each of the first level blocks (so that this model can also be realized as a 48 block standard SBM).
Further, the blocks of the first level belong to one of three motifs $B_1,\ B_2$, or $B_3$, with constant cross-community connection probability $p=0.01$ for the first level. 
Formally, we are sampling from the following 2-level HSBM:
\begin{itemize}
	\item[i.] There are $K_1=16$ blocks in the first level of the hierarchy, and the size of the blocks is i.i.d. $10*\lfloor20+50*Unif(0,1)\rfloor$; note that this differs slightly from the random block assignment HSBM defined previously.
	\item[ii.] The motifs $\mathbf{B}_i\in\mathbb{R}^{3\times 3}$ for $i=1,2,3$ are i.i.d. samples satisfying  
	\begin{align*}
	    &\mathbf{B}_i=X_i^TX_i\text{ where }X_i\in\mathbb{R}^{3\times 3}\\
	    &\text{ satisfies }X_i[\cdot,j]\stackrel{i.i.d.}{\sim} Dirichlet(1,1,1); 
	\end{align*}
of course, this distribution is artifical, but it provides a maximum entropy sampling of the X's which provides a difficult testing setting for our algorithm.
	\item[iii.] For each $j\in[16]\setminus\{9\}$, we have $B_2^{j}=\mathbf{B}_i$ independently with probability $1/3$ for $i=1,2,3$.
	$B_2^{9}$ is set to equal $B_2^{1}$;
	\item[iv.]  In the second level of the hierarchy ($K_2^j=3$ for all $j$), the block sizes are defined via 
	(where $\lfloor\cdot\rceil_1$ rounds the number to the nearest tenth)
	\begin{align*}
	    |b^{(2)}_{(j,i)}|&=|\{v\in V|  b^{(1)}(v)=j,b^{(2)}(v)=i\}|\\
	    &\distas{i.i.d.}|b^{(1)}_{j}|\lfloor Dirichlet\left(\omega \right)\rceil_{1}(i), i< 3
	\end{align*}
	where the entries of $\omega=(\omega_1,\omega_2,\omega_2)$ are \[\omega_i\distas{i.i.d.} Unif(2,10).\]
	Lastly, we set
	\begin{align*}
	|b^{(2)}_{(j,3)}|&=|\{v\in V| b^{(1)}(v)=j,b^{(2)}(v)=3\}\\
	&=|b^{(1)}_{j}|-|b^{(2)}_{(j,i)}|-|b^{(2)}_{(j,i)}|
	\end{align*}
\end{itemize}
This particular parameterization is chosen to produce different motifs that stress our HSGN framework by producing motifs that are very similar, and motifs where the block structure is extremely subtle (i.e., close to flat), and cases where both these problems happen, as it draws from a distribution in which it is not uncommon that the motifs turn out to be extremely similar and/or very flat.
\begin{figure*}[t!]
	\centering
	\begin{tabular}{cc}
	\includegraphics[width=7cm]{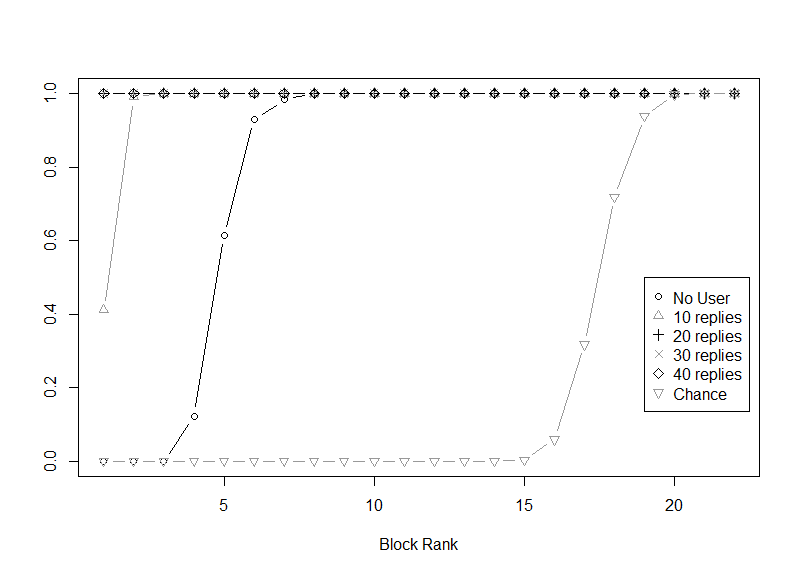}&
	\includegraphics[width=7cm]{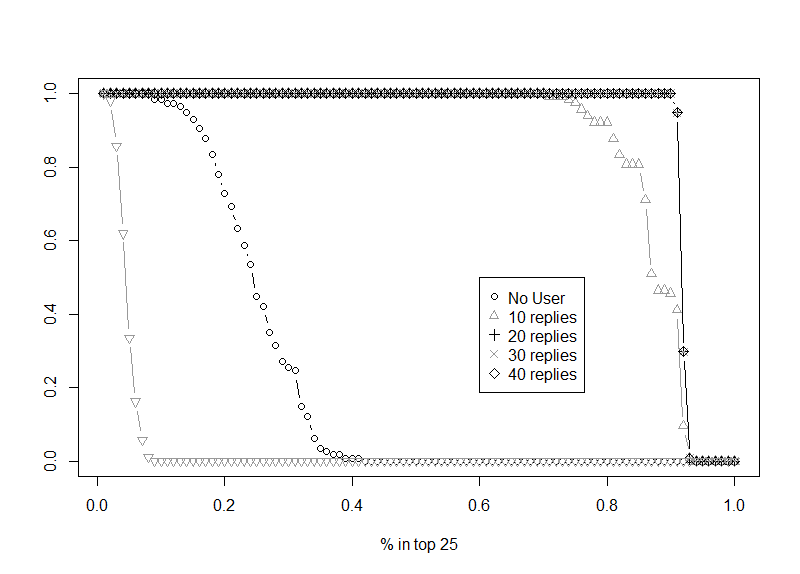}
	\end{tabular}
	\caption{With perfect knowledge of the block structure of the brain (the division into 71 regions provided by the data) and using Method 1, (left) we plot on the $y$-axis the proportion of subjects (averaged across all blocks considered as the block-of-interest in the left hemisphere) that ranked the correct block in the right hemisphere the $n$-th place (x-axis); 
		(right) we plot on the $y$-axis the the proportion of subjects (averaged across all blocks considered as the block-of-interest in the left hemisphere) for which we find the total proportion of the block of interest in the top 25 vertices nominated (x-axis).
	}
	\label{fig:blockleveltrue}
\end{figure*}

The block structure was inferred by clustering via Gaussian mixture modeling after embedding, utilizing the {\bf R} package {\bf Mclust} \cite{fraley1999mclust} to cluster the graph we are nominating from into 8 sub-graphs. 
After that, we use two different methods to infer similarity.
The first approximates $\boldsymbol{\Delta}$ via the value of the non-parametric test statistic of \cite{tang14:_nonpar} computed between re-embeddings of each inferred community (as in \cite{lyzinski2015community}); simulation results are shown in Figures \ref{fig:sim1-1} and \ref{fig:sim1-2}.
We will call this \emph{Method 1} in the sequel. 
The second approximates $\boldsymbol{\Delta}$ between communities $i$ and $j$ via the scaled graph matching pseudo-distance \cite{fishkind2019alignment},
$$
1-\frac{\min_{P\in\Pi(n)}\|A_i-PA_jP^T\|_F^2}{\frac{1}{n!}\sum_{P\in\Pi(n)} \|A_i-PA_jP^T\|_F^2 }
$$ 
where either $A_i$ (the induced subgraph of community $i$) or $A_j$ has been appropriately padded as in \cite{FAP});
results are shown in Figures \ref{fig:sim2-1} and \ref{fig:sim2-2}.
We will call this \emph{Method 2} in the sequel.

The user replies are those of an oracle. 
However, as discussed previously since we do perfectly recover the true network partition (rather than using an estimate of the true network partition), our user could still end up being errorful. In particular,
the user could be asked to rank vertices that are not in the subgraph of interest.
This exemplifies the behavior characterized by $iii$ of  Theorem \ref{thm:losspq}, albeit with a different $q$ for each of the blocks. 
Another significant difference in the scheme of the simulation is that the nomination process takes the first affirmative reply as the correct community and returns the others in their order as is.

\begin{figure*}[t!]
	\centering
	\begin{tabular}{cc}
		\includegraphics[width=7cm]{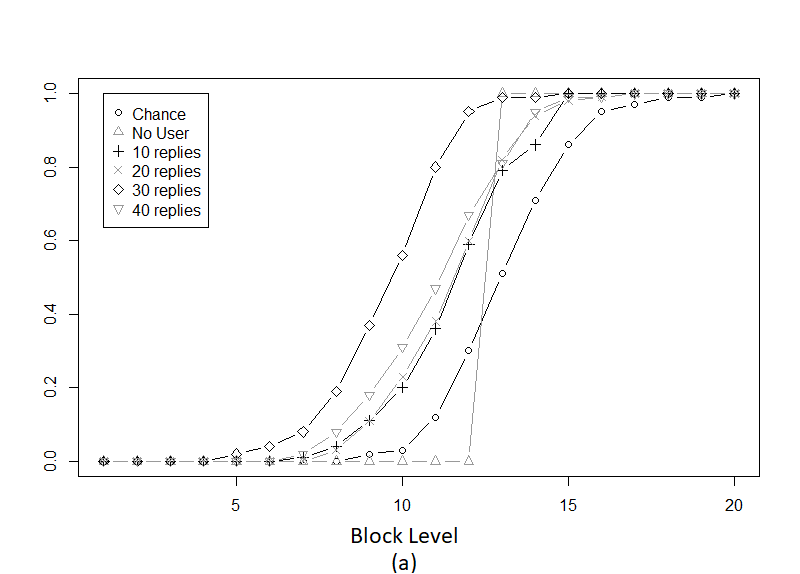}&
		\includegraphics[width=7cm]{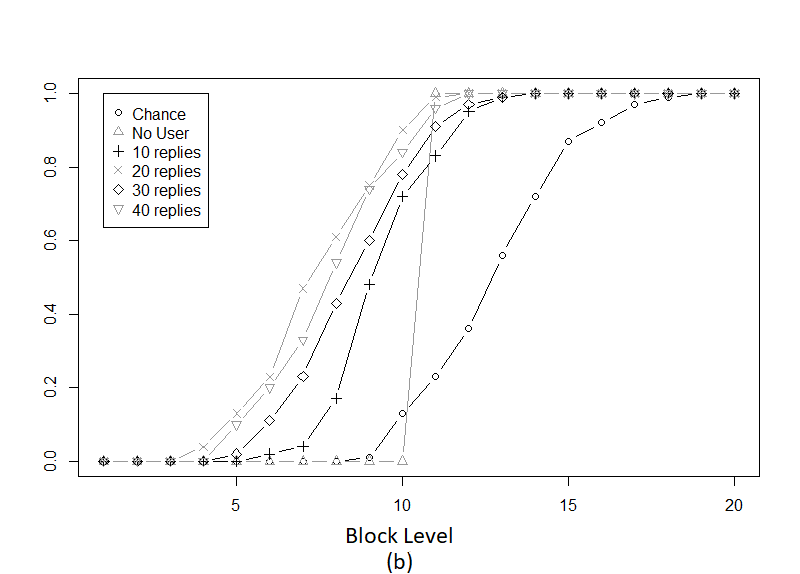}\\
		\includegraphics[width=7cm]{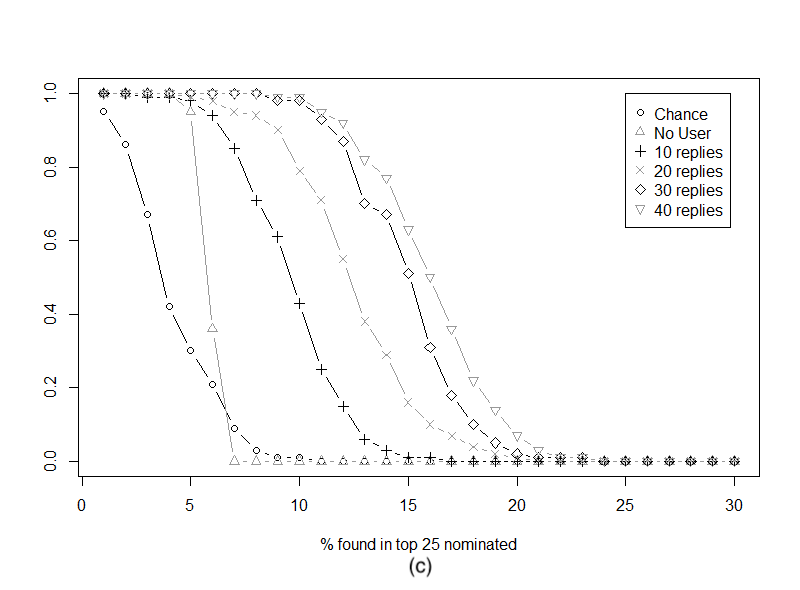}&
		\includegraphics[width=7cm]{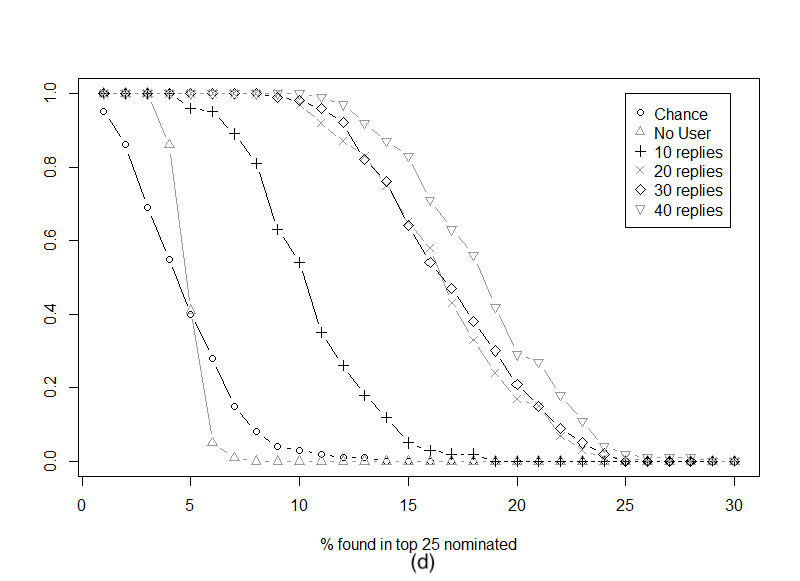}
	\end{tabular}
	\caption{Clustering the graph using Gaussian Mixture Modeling on the embedded network, in the left two panels, we use Method 1 to estimate $\boldsymbol{\Delta}$, while the right two panels we use Method 2. 
		In the top row, on the $y$-axis we plot the proportion of subjects (averaged across all blocks considered as the block-of-interest in the left hemisphere) that ranked the correct block in the right hemisphere the at worst $x$-th place (x-axis). 
		In the bottom row, on the $y$-axis we plot the the proportion of subjects for which we find the total proportion of the block of interest in the top 25 vertices nominated (x-axis).}
	\label{fig:blocklevelgoodvsbad}
\end{figure*}

\subsubsection{Results \& Discussion}

In this section we will go over the results of the experiments we described. 
The aim is to bolster the claims in the theoretical results with experimental results, as such we will have a special focus on the features that are iconic from the theory. 
First we start with Method 1, the experiment that uses the test statistic form \cite{tang14:_nonpar} to estimate the dissimilarity function.

In Figure \ref{fig:sim1-1} we can see that the user training provides an improvement over the original algorithm (``No user''), as suggested by Theorem \ref{thm:losspq}, with more training often yielding superior performance.
Moreover, especially for small $x$-values, the the algorithm outperforms a chance ranking of the obtained blocks.
Interestingly, we see that more user-supervision is not always better.
Indeed, as mentioned previously, in our regime of lossy block recovery 
additional supervision can be deleterious. 
Figure \ref{fig:sim1-2} elucidates this finding further from the perspective of the distribution density of the correct block in each position. 
We notice that as user-supervision increases the distribution are shifted towards putting the correct block in the first position. 
We also notice the trend predicted by Figure \ref{fig:numericL2}, as we increase $h$ (the nomination error level) the ideal value of $t$ (user-provided replies) is decreasing, and less supervision is preferred.
We see this, as the curves with less supervision overtake those of higher supervision as we plot $1-L_n(\phi,B^*)$. However we note that here the value of $q$ is not constant and hence the minimizers of the loss function do not not adhere to the figure exactly.
In Figures \ref{fig:sim2-1} and \ref{fig:sim2-2} we see similar trends using Method 2.
Here the main contribution to the error of the methods is the misclustering of the vertices in the initial GMM step, and both 
dissimilarity estimates provide good estimates of the the latent $\boldsymbol{\Delta}$.
In the real data example considered below, there is a more pronounced differentiation across methods. 

\subsection{Subgraph nomination in BNU1 connectomes}
\label{expr}
Data acquisition resources and limitations can be eased through automation. 
For example, in the study of human connectomes, one of the more labour intensive steps, is classifying different parts of the brain. One such classification task is finding corresponding regions of interest across hemispheres in a human connectome. 
This task is time and human resource intensive when done manually, and automation is essential for achieving a high throughput \cite{gray2012magnetic}.
To illustrate our HSN methodology further, we will consider then the task of nominating brain regions across hemispheres in the DT-MRI derived brain networks in the BNU1 database \cite{zuo2014open}.
The spatial image of the brain includes  information about its structure; in other words, the neural fibers that run through different areas of the brain. 
A graph of the subject's brain is then created with the following definitions: nodes or vertices correspond to different spatial units (usually the brain is divided into cubic sections called volumetric pixels, or voxels), edges are weighted based on the amount of neural fibers that run through two regions. 
One may hypothesize that there is a sense of structural symmetry between paired regions across hemispheres, and discovery of this pairing is the inference task for this HSN application.

Here, we apply the user aided subgraph nomination scheme to a connectomic dataset. 
The data we apply this to is a set of 114 neuronal networks---two repeated scans for each of 57 human subjects---derived from the BNU1 connectome dataset \cite{zuo2014open}.
Each brain scan sections the brain into one thousand voxels---or volumetric pixels---with weighted edges between them indicating the amount of neuronal batches that are shared, indicating connectedness in a structural but not necessarily functional sense.
Additionally each voxels contains information on it detailing its membership to the left or right hemispheres, membership to one of the 71 neuronal regions  \cite{desikan2006automated}, whether it is grey or white matter, and its XYZ coordinates in the partially registered scan.
The relative size of the networks is then $\approx 1000$ vertices, with pre-defined neuronal regions varying in size between 2 and 100 vertices. 
As in the simulation, although the distributional properties in paired regions across hemispheres is often similar (i.e. similar motifs), there are no guarantees that the numbers of vertices that make up each region are equal. 
Errors from this (and misclustering) are magnified in smaller subgraphs, especially since there is less information about the motif in the small subgraph setting. 
The aforementioned sources of errors are mitigated by only considering paired regions that have at least 10 vertices in each hemisphere and using similarity metrics (Methods 1 and 2) that do not depend on the size of a region. 

\begin{figure*}[t!]
	\centering
	\begin{tabular}{cc}
	\includegraphics[width=7cm]{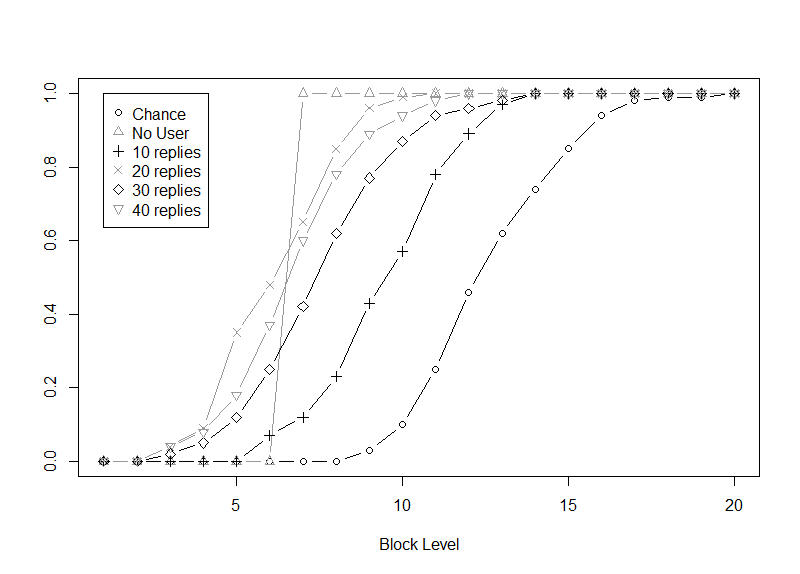}&
	\includegraphics[width=7cm]{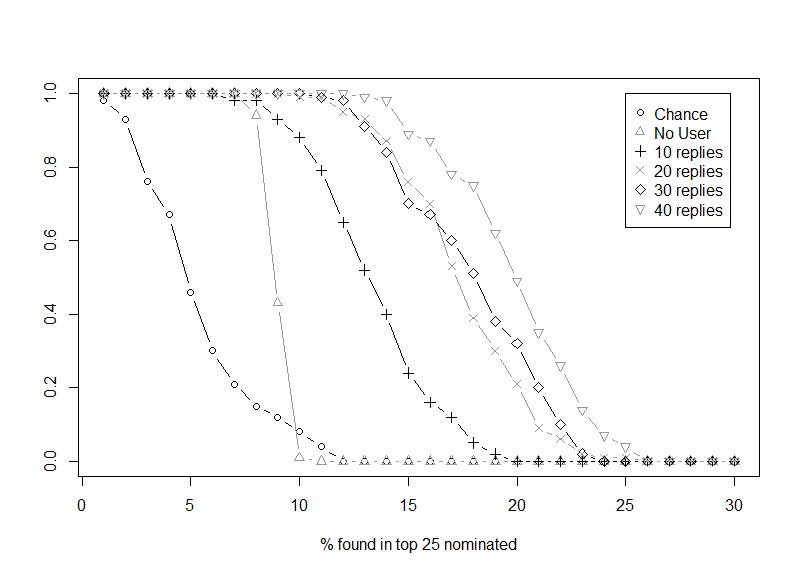}
	\end{tabular}
	\caption{Clustering the embedded graphs using Gaussian Mixture Modeling incorporating the XYZ  coordinate data and using Method 2 to estimate $\boldsymbol{\Delta}$.
		In the left panel, on the $y$-axis we plot the proportion of subjects (averaged across all blocks considered as the block-of-interest in the left hemisphere) that ranked the correct block in the right hemisphere the at worst $x$-th place (x-axis). 
		In the right panel, on the $y$-axis we plot the the proportion of subjects (averaged across all blocks considered as the block-of-interest in the left hemisphere) for which we find the total proportion of the block of interest in the top 25 vertices nominated (x-axis).}
	\label{fig:blocklevelXYZ}
\end{figure*}

Our procedure can then be described as follows, considering each pair of sufficiently large matched regions ($\geq 10$ vertices in each hemisphere) separately, first we use the information given by spectrally embedding the adjacency matrix of the connectome (and the collected spatial coordinate data in Figure 
\ref{fig:blocklevelXYZ}) to infer the subgraph structure in the right hemisphere via clustering by GMM \cite{mclust};
using the region-of-interest in the left hemisphere as our training, next we use Methods 1 and 2 to rank above to provide different estimates of  $\boldsymbol{\Delta}$.
We then supply a user-in-the-loop with a vertex (or a few vertices) from each of the top $k$ communities to re-rank the nomination list. 
We plot the performance of our algorithms with perfect knowledge of the block structure of the brain (the division into 71 regions provided by the data) and using Method 1 in Figure \ref{fig:blockleveltrue}.
We plot:
\begin{itemize}
	\item (Left) on the $y$-axis the proportion of subjects (averaged across all blocks considered as the block-of-interest in the left hemisphere) that ranked the correct block in the right hemisphere in the at worst $x$-th place (x-axis). 
	\item (Right) on the $y$-axis the the proportion of subjects for which we find the total proportion of the block of interest in the top 25 vertices nominated (x-axis).
\end{itemize}
Note that an artifact of the metric in panel (b) is that we may never find a complete block that is of a size larger than 25 this is why we see a drop near 90$\%$.
As ideal performance in the left and right panels corresponds to the function $f(x)\equiv 1$, this figure enforces the intuition that, given high fidelity clusters, our ranking procedure is significantly better than chance and benefits strongly from user-in-the-loop supervision.

In Figure \ref{fig:blocklevelgoodvsbad}, in the setting of potentially errorful clusters (obtained via \texttt{Mclust} applied to the adjacency spectral embeddings of the brain networks, we plot (similar to in Figure \ref{fig:blockleveltrue}), in the top panels the proportion whose match is found in the top $x$ (similar to the left panel of Figure \ref{fig:blocklevelgoodvsbad}) and in the bottom panels the percent who had the proportion of the top 25 vertices come from the block of interest (similar to the right panel of Figure \ref{fig:blocklevelgoodvsbad}).
The left two panels use Method 1 to estimate $\boldsymbol{\Delta}$, while the right two panels use Method 2.
In light of perfect performance corresponding the the constant 1 function in all cases, we see here that Method 2 achieves better performance (especially for large values on the $x$-axis in each figure) than Method 1 both when making use of the user-supervision and in the no-user setting. 
Here, we see the general trend that the methods perform poorly without a user-in-the-loop (due to the error in the clustering; see Figure \ref{fig:blocklevelXYZ}), but that the method can efficiently make use of the user to achieve better performance.  
As was the case previously, the clustering is errorful, and on average this can have a deleterious effect on the user-supervision (top row).
However, in the bottom row we see that \emph{sample-wise} the supervision is monotonically beneficial.

We repeat the above experiment using Method 2 and incorporating the XYZ coordinates of each voxel into the clustering (by appending the features onto the spectral graph embeddings), and plot the results in Figure \ref{fig:blocklevelXYZ}.
The increased clustering fidelity achieved by incorporating the vertex features manifests itself here by the performance gain achieved in the no-user setting when compared with Figure \ref{fig:blocklevelgoodvsbad}.
As a result of this, better performance is achieved across all settings here (again when compared with Figure \ref{fig:blocklevelgoodvsbad}).
While the clustering here is still errorful, and on average this can have a deleterious effect on the user-supervision, the right panel again shows that sample-wise the supervision is monotonically beneficial.
This means that a user can successfully aid in finding larger portions of the the block of interest. Therefore, we can use these vertices in the input to further investigate and find the rest of the block, by searching in the $k$-nearest neighbours of these vertices.

\subsection{Subgraph nomination on the Reddit social network}

Social network moderation is another area where the biggest limitation is the human resources required to analyze the prohibitively large amount of data. One problem that appears in Reddit, is finding subreddits for which the name has changed. This can be needed to track a previously banned subreddit whose users create new accounts and a new subreddit. Further, since this experiment depends only on structure, which is informed by user interaction patterns, it could be used for recommender systems, or finding suspicious behaviour in other subreddits. The first is under the assumptions that people who communicate in the same way may enjoy similar content. The second can be done by giving examples of subreddits which show suspicious behavior and looking at other subreddits that are embedded close to those subreddits. The idea behind the automation here is to distill the data so an expert moderator can take a look at the smaller data.

The data was collected by \cite{baumgartner2020pushshift}. The data includes information about subreddits (S), their threads (T), comments (C) on these threads, users (U) that posted the thread or comment, and flairs (F) associated with the subreddits. These form our node types, the focus here is on structure so we do not maintain node types. We also dispose of the users, as they will form a leaf coming out of each comment and thread. Each of these comes with a multitude of features, notably parent comment/thread, and containing subreddit. All features are discarded for a structure based approach, however we use the relations above to create the graph. Our edges represent $S\leftrightarrow T$, $T\leftrightarrow C$ if $T$ is not parent of $C$, else $T\leftarrow C$, $C_1\leftarrow C_2$ if $C_1$ is parent of $C_2$, and $S\leftrightarrow F$. A graph is constructed for the dataset, then the subgraph for each subreddit is retrieved. Using the post-time date data each month can be subset to 4 week time periods. The data is subset to subreddits of size 1000 to 2000 nodes.

The Graph Matching Network from \cite{li2019graph}, a Graph Neural Network approach for comparison at the graph level, is used as the graph similarity (or dissimilarity) function. The neural network embeds graphs into a space where the closer two graphs are to one another the more similar they are, the normalized embeddings provide a similarity metric. At this point, normally the top $h$ recommendations are passed to an expert to analyse. The expert would analyze the subreddits to make a decision. Here, we can insert a user in the loop, the user would consider a node (or small collection of nodes) in the graph instead of the entire graph. Here, where we are looking to recover the subreddits in time, the node-level user defines a correct match if the node is in the subreddit of interest. 
Training is done on the first 2 weeks of the reddit data, and the remaining two weeks are used to test the neural network's performance and implement the user. Note that the user input can be used as a node feature and implemented during the training as well; we do not pursue that here.

\begin{figure*}[t!]
	\centering
	\begin{tabular}{cc}
	\includegraphics[width=7cm]{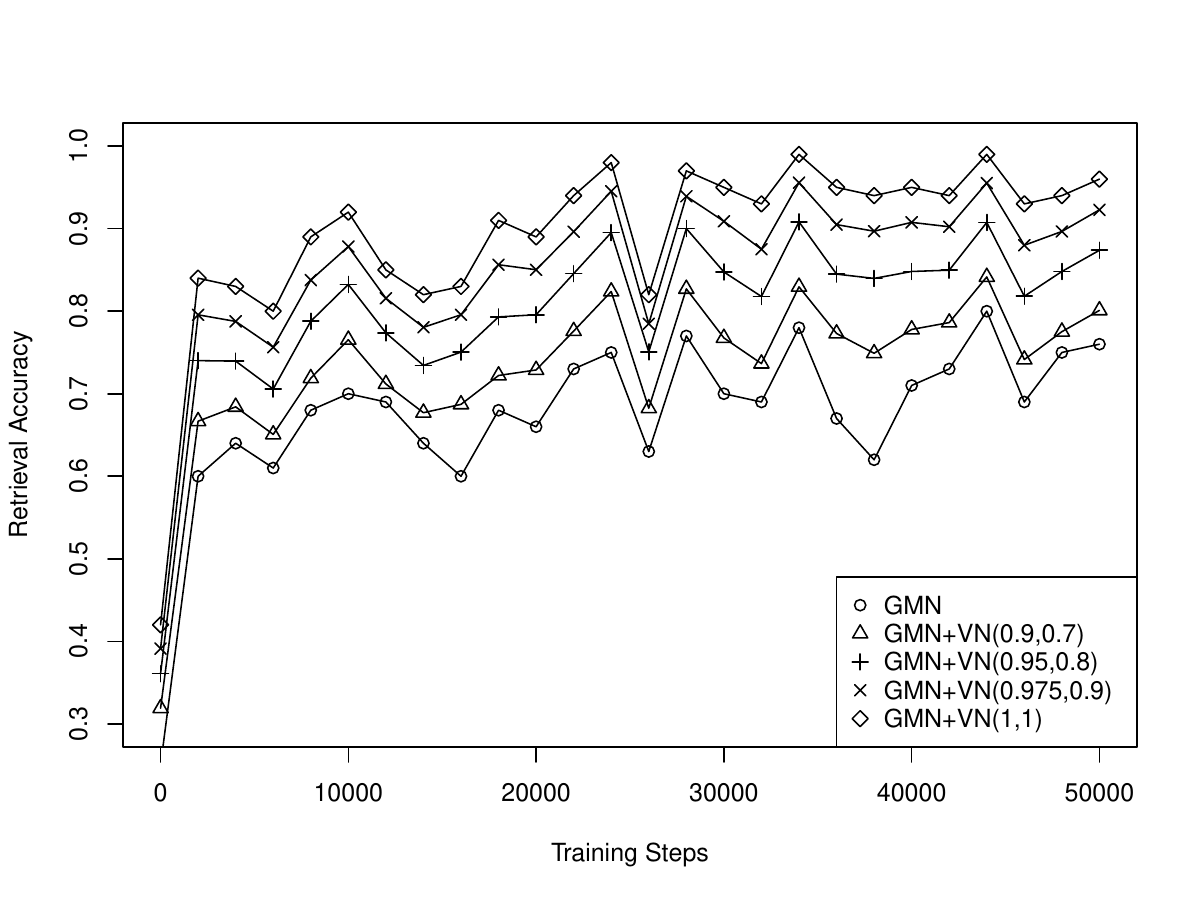}&
        \includegraphics[width=7cm]{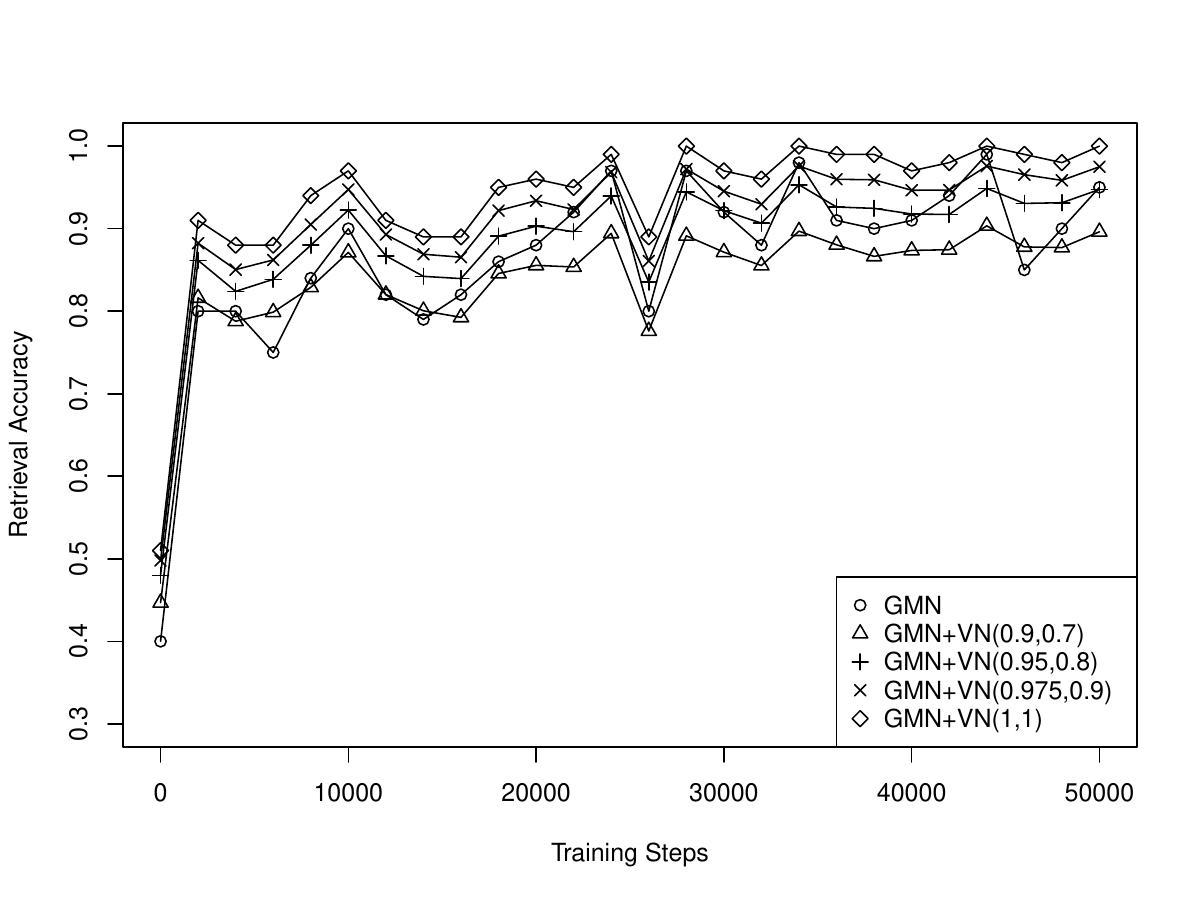}
	\end{tabular}
	\caption{Clustering the graphs using their subreddit membership and using GMN to estimate $\boldsymbol{\Delta}$.
		The $y$-axis we plot the proportion of subreddits (averaged across 100 runs of the VN user) that ranked the correct subreddit in the second time period $x$-axis is the training time .
        In the left panel, we look at the top 2 nominations.
	In the right panel, we look at the top 8 nominations.}
	\label{fig:GMN+VN}.
\end{figure*}

After the training is complete and the tests are performed, the top 8 subgraphs are nominated, and we simulate VN users-in-the-loop with different values for $\theta$ and $\gamma$. The result is displayed in Figure \ref{fig:GMN+VN}, where the x-axis is the number of GMN training steps and the y-axis represents the retrieval accuracy. Accuracy here is the percentage of subgraphs for which the correct community is nominated in the top $\ell$ graphs. Denoting the user $U_8$ with parameters $(\theta,\gamma)$ with $VN(\theta,1-\gamma)$ (recalling that $\theta$ and $1-\gamma$ are the probabilities of correct user input for true positive and true negative nodes resp.), the left plot is the accuracy for $\ell=2$ and on the right is for $\ell=8$.

Notice in Figure \ref{fig:GMN+VN} the addition of the user-in-the-loop helps achieve better results especially at the lower level $\ell=2$. Also, we can achieve desirable results with less training, even with a faulty VN user. The performance at $\ell=2$ suggests the user can lower the costs of verification, as an expert user needs to look at less subgraphs to determine whether the nomination is successful.
Moreover, this suggests that we can automate the VN user-in-the-loop by using node features to make decisions, and train the GMN for a shorter amount of time to reduce the computational cost. This is critical since the GMN computational cost grows quadratically with the total number of nodes in the graphs.

\section{Conclusions and Future Direction}
In this paper we have introduced a formal, and versatile framework for the novel subgraph nomination inference task, with special emphasis paid to the utility of users-in-the-loop in the context of subgraph nomination. 
Subgraph nomination is an important tool for structural queries within and across networks, and we demonstrated its utility in both real and synthetic data examples.
In addition, the relatively simple users-in-the-loop formulation outlined herein can easily be lifted to more complex cases including supervision done over several iterations, and can be simply extended to the case of multiple users.
The theory and experiments included herein highlight the important role that user supervision can play in effective information retrieval. 

Further, this paper shows the importance of analysing the original algorithm and the validity of user input before including them. 
For example, if the inferred clustering is incorrect in a hierarchical subgraph nomination framework, in the sense that the correct clusters' vertices are evenly spread among the inferred blocks, then our resources should be focused on collecting relevant data that would improve the inferred blocking structure, as even oracle user supervision in this case could be detrimental to the performance. 
In our approach to subgraph nomination, inferring high-fidelity clusters is essential, and we have demonstrated that clustering can be improved by incorporating informative vertex features.
Exploring this further, understanding the information theoretic gain of features on our nomination performance as in \cite{levin2020role}, is an open area of research that we are actively pursuing.

The validity of the user input itself is also important to consider when thinking about adding a user-in-the-loop. 
While we have theoretically demonstrated the possibility of increased performance even with an errorful user, in practice the situation is significantly more nuanced.
The user errors, rather than being uniformly random, may be systematic or deterministic.
For example an adversarial attacks could manipulate the results of the search using a user bot attack.
More nuanced users-in-the-loop demand more nuanced theory to understand their broad effects.
As an example, in our nomination regime an adversary can target the obtained cluster labels for contamination.
This would immediately mitigate the benefit of a user-in-the-loop, as the benefit of the user decays as the cluster fidelity worsens.
More robust users would be, most likely, more costly and a cost-benefit optimization analysis would be needed in these more nuanced setting to tease out the positive (or negative) impact of incorporating the user.

Another avenue that has not been investigated explicitly but for which the framework is still sufficient is the situation with multiple blocks of interest. 
For that we need to adjust the definition of the VN-user to accommodate for different values of $p$ for each block of interest. 
Then, after some necessary adjustment to the loss function, one may obtain results generalized to the case of multiple blocks of interest in the simple user-in-the-loop setting. 
On the other hand, one may want to extend the results of Theorem \ref{thm:losspq} to more nuanced users. 

\textcolor{black}{
We close with a discussion of scalability and computational complexity. For implementing the core subgraph nomination routine, the computational burden is in graph partitioning and subgraph comparison, both of which can be implemented efficiently; e.g., any number of scalable clustering routines for graph division, and fast graph similarity algorithms (e.g., \cite{sim}).
For the VN user-in-the-loop it will greatly depend on the setting, essentially we randomly retrieve an element from each of the top $h$ communities. We then apply the user, if the user is an expert then the cost is not only computational (if the expert needs to perform some analysis on the node's feature for example) but also in human resources, which is beyond the scope of this paper, but intuitively scalability is an issue. If an automated user is used (e.g., one who is trained on the node features alongside the algorithm) then the cost is that of training and $h$ times the cost of retrieving the model's fit on the selected nodes, here scalability is less of an issue. 
}
\vspace{2mm}

\noindent{\bf Acknowledgement}
This material is based on research sponsored by the Air Force Research Laboratory and DARPA under agreement number FA8750-20-2-1001.
The U.S. Government is authorized
to reproduce and distribute reprints for Governmental purposes notwithstanding any
copyright notation thereon. The views and conclusions contained herein are those of
the authors and should not be interpreted as necessarily representing the official policies
or endorsements, either expressed or implied, of the Air Force Research Laboratory and
DARPA or the U.S. Government. The authors also gratefully acknowledge the support
of NIH grant BRAIN U01-NS108637 and the JHU HLTCOE.

\section*{APPENDIX I: Proof of Theorem \ref{thm:losspq}}
\label{proof17}

For $i\leq j\leq c$, let the event $A_{[i,j]}$ be the event that the unknown subgraph of interest in $g_2$ has rank in $[i,j]$.
Let $E_h$ be the event that after user-in-the-loop supervision, the subgraph of interest has rank strictly greater than $h$.
\vspace{2mm}

\noindent\emph{Part I:} With an oracle user-in-the-loop, we have that
\begin{align*}
	\mathbb{P}(E_h)&=\underbrace{\mathbb{P}(E_h| A_{[1,h]})}_{=0} \mathbb{P}(A_{[1,h]}) \\
	&~~~+\underbrace{\mathbb{P}(E_h| A_{[h+1,h+t]})}_{=0} \mathbb{P}(A_{[h+1,h+t]})\\
	&~~~+\underbrace{\mathbb{P}(E_h| A_{[h+t+1,m_k]})}_{=1} \mathbb{P}(A_{[h+t+1,m_k]})\\
	&=\mathbb{P}(A_{[h+t+1,m_k]})
\end{align*}
If $h+t\leq c$, then $\mathbb{P}(A_{[h+t+1,k]})=1-\frac{h+t}{c}$, else it is $0$.  Hence, $\mathbb{P}(E_h)=\max(0,1-\frac{h+t}{c})$ as desired.
\vspace{2mm}

\noindent\emph{Part II:} When the user has probability $p$ of misidentifying the vertex from the subgraph of interest as non-interesting, then when $h+t\leq c$, 
\begin{align*}
	\mathbb{P}(E_h)&=\underbrace{\mathbb{P}(E_h| A_{[1,t]})}_{=p} \mathbb{P}(A_{[1,t]})\\
	&~~~+ \underbrace{\mathbb{P}(E_h| A_{[t+1,h+t]})}_{=0} \mathbb{P}(A_{[t+1,h+t]})\\
	&~~~+\underbrace{\mathbb{P}(E_h| A_{[h+t+1,m_k]})}_{=1} \mathbb{P}(A_{[h+t+1,m_k]})\\
	&=p\mathbb{P}(A_{[1,t]})+ \mathbb{P}(A_{[h+t+1,m_k]})\\
	&=\frac{pt}{c}+ 1-\frac{h+t}{c}=1-(1-p)\frac{t }{c}-\frac{h}{c}.
\end{align*}
When $m_k> h+t> c$, the above probability is simply $\frac{pt}{c}$.

\vspace{2mm}

\noindent\emph{Part III:} 
When the user has probability $p$ of misidentifying the vertex from the subgraph of interest as non-interesting and probability $q$ of misidentifying a non-interesting vertex as interesting, then when $h+t\leq c$ and $t\leq h$, 
\begin{align*}
	\mathbb{P}(E_h)&=\underbrace{\mathbb{P}(E_h| A_{[1,t]})}_{=p} \mathbb{P}(A_{[1,t]}) \\
	&~~~+ \underbrace{\mathbb{P}(E_h| A_{[t+1,h]})}_{=0} \mathbb{P}(A_{[t+1,h]})\\
	&~~~+\sum_{i=1}^{t} \left[\mathbb{P}(E_h| A_{[h+i,h+i]}) \underbrace{\mathbb{P}(A_{[h+i,h+i]})}_{=\frac{1}{c}}\right.\\
	&~~~\left.+\underbrace{\mathbb{P}(E_h| A_{[h+t+1,m_k]})}_{=1} \mathbb{P}(A_{[h+t+1,m_k]})\right]\\
	&=\frac{pt}{c}+ \sum_{i=1}^{t}\frac{1}{c}\sum_{j=0}^{i-1}\binom{t}{j}(1-q)^jq^{t-j}\\
	&~~~+1-\frac{h+t}{c}\\
	&=\frac{pt}{c}+ \sum_{j=0}^{t-1}\frac{t-j}{c}\binom{t}{j}(1-q)^jq^{t-j}\\
	&~~~+1-\frac{h+t}{c}\\
	&=\frac{pt}{c}+ \sum_{j=0}^{t}\frac{t-j}{c}\binom{t}{j}(1-q)^jq^{t-j}\\
	&~~~+1-\frac{h+t}{c}\\
	&=\frac{pt}{c}+\frac{qt}{c}+1-\frac{h+t}{c}.
\end{align*}
When $h+t> c$ and $t\leq h$, 
\begin{align*}
	\mathbb{P}(E_h)&=\underbrace{\mathbb{P}(E_h| A_{[1,t]})}_{=p} \mathbb{P}(A_{[1,t]})\\
	&~~~+\underbrace{\mathbb{P}(E_h| A_{[t+1,h]})}_{=0} \mathbb{P}(A_{[t+1,h]})\\
	&~~~+\sum_{i=1}^{c-h} \mathbb{P}(E_h| A_{[h+i,h+i]}) \underbrace{\mathbb{P}(A_{[h+i,h+i]})}_{=\frac{1}{c}}\\
	&~~~+\mathbb{P}(E_h| A_{[c+1,m_k]}) \underbrace{\mathbb{P}(A_{[c+1,m_k]})}_{=0}\\
	&=\frac{pt}{c}+ \frac{1}{c}\sum_{i=1}^{c-h}F(i-1;t,1-q)
\end{align*}
When $t>h$, and $h+t\leq c$, then
\begin{align*}
	\mathbb{P}(E_h)&=\underbrace{\mathbb{P}(E_h| A_{[1,h]})}_{=p} \mathbb{P}(A_{[1,h]}) \\
	&~~~+ \sum_{i=1}^{t-h} \mathbb{P}(E_h| A_{[h+i,h+i]}) \underbrace{\mathbb{P}(A_{[h+i,h+i]})}_{=\frac{1}{c}}\\
	&\hspace{4mm}+\sum_{i=t-h+1}^{t} \mathbb{P}(E_h| A_{[h+i,h+i]}) \underbrace{\mathbb{P}(A_{[h+i,h+i]})}_{=\frac{1}{c}}\\
	&~~~+\underbrace{\mathbb{P}(E_h| A_{[h+t+1,m_k]})}_{=1} \mathbb{P}(A_{[h+t+1,m_k]})\\
	&=\frac{ph}{c} \sum_{i=1}^{t-h}\frac{1}{c}\Bigg (p\\
	&\hspace{8mm}\left.+(1-p)\sum_{j=0}^{i-1}\binom{h+i-1}{j}(1-q)^jq^{h+i-1-j}\right)\\
	&~~~+\sum_{i=t-h+1}^{t}\frac{1}{c}\sum_{j=0}^{i-1}\binom{t}{j}(1-q)^jq^{t-j}
	+1-\frac{h+t}{c}\\
	&= 1-\frac{h}{c}-\frac{(1-p)t}{c}\\
	&\hspace{4mm}+\frac{1-p}{c}\sum_{i=1}^{t-h}F(i-1;h+i-1,1-q)\\
	&\hspace{4mm}+\frac{1}{c}\sum_{i=t-h+1}^tF(i-1;t,1-q)
\end{align*}
When $t>h$, and $h+t> c$, then (assuming $c\geq t,h$)
\begin{align*}
	\mathbb{P}(E_h)&=\underbrace{\mathbb{P}(E_h| A_{[1,h]})}_{=p} \mathbb{P}(A_{[1,h]}) \\
	&\hspace{4mm}+\sum_{i=1}^{t-h} \mathbb{P}(E_h| A_{[h+i,h+i]}) \underbrace{\mathbb{P}(A_{[h+i,h+i]})}_{=\frac{1}{c}}\\
	&\hspace{4mm}+\sum_{i=t-h+1}^{c-h} \mathbb{P}(E_h| A_{[h+i,h+i]}) \underbrace{\mathbb{P}(A_{[h+i,h+i]})}_{=\frac{1}{c}}\\
	&\hspace{4mm}+\mathbb{P}(E_h| A_{[c+1,m_k]}) \underbrace{\mathbb{P}(A_{[c+1,m_k]})}_{=0}\\
	&= \frac{pt}{c}+\frac{1-p}{c}\sum_{i=1}^{t-h}F(i-1;h+i-1,1-q)\\
	&\hspace{4mm}+\frac{1}{c}\sum_{i=t-h+1}^{c-h} F(i-1;t,1-q)
\end{align*}

\bibliographystyle{plain}

\bibliography{biblio}

\begin{thebibliography}{10}

\bibitem{agterberg2019vertex}
J.~Agterberg, Y.~Park, J.~Larson, C.~White, C.~E. Priebe, and V.~Lyzinski.
\newblock Vertex nomination, consistent estimation, and adversarial
  modification.
\newblock {\em Electronic Journal of Statistics}, 14(2):3230--3267, 2020.

\bibitem{sg1}
N.~Alon, R.~Yuster, and U.~Zwick.
\newblock Color-coding.
\newblock {\em Journal of the ACM (JACM)}, 42(4):844--856, 1995.

\bibitem{amershi2014power}
S.~Amershi, M.~Cakmak, W.~B. Knox, and T.~Kulesza.
\newblock Power to the people: The role of humans in interactive machine
  learning.
\newblock {\em Ai Magazine}, 35(4):105--120, 2014.

\bibitem{angles2016foundations}
Renzo Angles, Marcelo Arenas, Pablo Barcelo, Aidan Hogan, Juan Reutter, and
  Domagoj Vrgoc.
\newblock Foundations of modern query languages for graph databases, 2016.

\bibitem{sim}
Y.~Bai, H.~Ding, S.~Bian, T.~Chen, Y.~Sun, and W.~Wang.
\newblock Simgnn: A neural network approach to fast graph similarity
  computation.
\newblock In {\em Proceedings of the Twelfth ACM International Conference on
  Web Search and Data Mining}, pages 384--392, 2019.

\bibitem{baumgartner2020pushshift}
Jason Baumgartner, Savvas Zannettou, Brian Keegan, Megan Squire, and Jeremy
  Blackburn.
\newblock The pushshift reddit dataset.
\newblock In {\em Proceedings of the international AAAI conference on web and
  social media}, volume~14, pages 830--839, 2020.

\bibitem{blondel2008fast}
V.~D. Blondel, J.-L. Guillaume, R.~Lambiotte, and E.~Lefebvre.
\newblock Fast unfolding of communities in large networks.
\newblock {\em Journal of statistical mechanics: theory and experiment},
  2008(10):P10008, 2008.

\bibitem{nsg2}
T.~Caelli and S.~Kosinov.
\newblock An eigenspace projection clustering method for inexact graph
  matching.
\newblock {\em IEEE transactions on pattern analysis and machine intelligence},
  26(4):515--519, 2004.

\bibitem{clauset08:_hierar}
A.~Clauset, C.~Moore, and M.~E.~J. Newman.
\newblock Hierarchical structure and the prediction of missing links in
  networks.
\newblock {\em Nature}, 453:98--101, 2008.

\bibitem{coppersmith2014vertex}
G.~Coppersmith.
\newblock Vertex nomination.
\newblock {\em Wiley Interdisciplinary Reviews: Computational Statistics},
  6(2):144--153, 2014.

\bibitem{sg3}
L.~P Cordella, P.~Foggia, C.~Sansone, and M.~Vento.
\newblock A (sub) graph isomorphism algorithm for matching large graphs.
\newblock {\em IEEE transactions on pattern analysis and machine intelligence},
  26(10):1367--1372, 2004.

\bibitem{cuts2}
D.~Delling, A.~V. Goldberg, I.~Razenshteyn, and R.~F. Werneck.
\newblock Graph partitioning with natural cuts.
\newblock In {\em 2011 IEEE International Parallel \& Distributed Processing
  Symposium}, pages 1135--1146. IEEE, 2011.

\bibitem{desikan2006automated}
R.~S. Desikan, F.~S{\'e}gonne, B.~Fischl, B.~T. Quinn, B.~C. Dickerson,
  D.~Blacker, R.~L. Buckner, A.~M. Dale, R.~P. Maguire, B.~T. Hyman, et~al.
\newblock An automated labeling system for subdividing the human cerebral
  cortex on mri scans into gyral based regions of interest.
\newblock {\em Neuroimage}, 31(3):968--980, 2006.

\bibitem{FAP}
D.~E. Fishkind, S.~Adali, H.~G. Patsolic, L.~Meng, D.~Singh, V.~Lyzinski, and
  C.~E. Priebe.
\newblock Seeded graph matching.
\newblock {\em Pattern recognition}, 87:203--215, 2019.

\bibitem{FisLyzPaoChePri2015}
D.~E. Fishkind, V.~Lyzinski, H.~Pao, L.~Chen, and C.~E. Priebe.
\newblock Vertex nomination schemes for membership prediction.
\newblock {\em The Annals of Applied Statistics}, 9(3):1510--1532, 2015.

\bibitem{Fishkind_2015}
D.~E. Fishkind, V.~Lyzinski, H.~Pao, L.~Chen, and C.~E. Priebe.
\newblock Vertex nomination schemes for membership prediction.
\newblock {\em The Annals of Applied Statistics}, 9(3):1510–1532, Sep 2015.

\bibitem{fishkind2019alignment}
D.~E. Fishkind, L.~Meng, A.~Sun, C.~E. Priebe, and V.~Lyzinski.
\newblock Alignment strength and correlation for graphs.
\newblock {\em Pattern Recognition Letters}, 125:295--302, 2019.

\bibitem{fraley1999mclust}
C.~Fraley and A.~E. Raftery.
\newblock Mclust: Software for model-based cluster analysis.
\newblock {\em Journal of Classification}, 16(2):297--306, 1999.

\bibitem{mclust}
C.~Fraley and A.~E. Raftery.
\newblock Mclust: Software for model-based cluster analysis.
\newblock {\em Journal of Classification}, 16(2):297--306, 1999.

\bibitem{frenay2013classification}
B.~Fr{\'e}nay and M.~Verleysen.
\newblock Classification in the presence of label noise: a survey.
\newblock {\em IEEE transactions on neural networks and learning systems},
  25(5):845--869, 2013.

\bibitem{gray2012magnetic}
W.~R. Gray, J.~A. Bogovic, J.~T. Vogelstein, B.~A. Landman, J.~L. Prince, and
  R.~J. Vogelstein.
\newblock Magnetic resonance connectome automated pipeline: an overview.
\newblock {\em Pulse, IEEE}, 3(2):42--48, 2012.

\bibitem{helm2020learning}
H.~S. Helm, A.~Basu, A.~Athreya, Y.~Park, J.~T. Vogelstein, M.~Winding,
  M.~Zlatic, A.~Cardona, P.~Bourke, J.~Larson, C.~White, and C.~E. Priebe.
\newblock Learning to rank via combining representations.
\newblock {\em arXiv preprint arXiv:2005.10700}, 2020.

\bibitem{sbm}
P.~W. Holland, K.~B. Laskey, and S.~Leinhardt.
\newblock Stochastic blockmodels: First steps.
\newblock {\em Social networks}, 5(2):109--137, 1983.

\bibitem{nsg1}
H.~Jin, X.~He, Y.~Wang, H.~Li, and A.~L. Bertozzi.
\newblock Noisy subgraph isomorphisms on multiplex networks.
\newblock In {\em 2019 IEEE International Conference on Big Data (Big Data)},
  pages 4899--4905. IEEE, 2019.

\bibitem{cuts}
V.~Kolmogorov and C.~Rother.
\newblock Minimizing nonsubmodular functions with graph cuts-a review.
\newblock {\em IEEE transactions on pattern analysis and machine intelligence},
  29(7):1274--1279, 2007.

\bibitem{levin2020role}
K.~Levin, C.~E. Priebe, and V.~Lyzinski.
\newblock On the role of features in vertex nomination: Content and context
  together are better (sometimes).
\newblock {\em arXiv preprint arXiv:2005.02151}, 2020.

\bibitem{bickel_hierarchical}
Tianxi Li, Lihua Lei, Sharmodeep Bhattacharyya, Purnamrita Sarkar, Peter~J
  Bickel, and Elizaveta Levina.
\newblock Hierarchical community detection by recursive partitioning.
\newblock {\em arXiv preprint arXiv:1810.01509}, 2018.

\bibitem{li2019graph}
Yujia Li, Chenjie Gu, Thomas Dullien, Oriol Vinyals, and Pushmeet Kohli.
\newblock Graph matching networks for learning the similarity of graph
  structured objects.
\newblock In {\em International conference on machine learning}, pages
  3835--3845. PMLR, 2019.

\bibitem{exemplar2}
M.~Lissandrini, D.~Mottin, T.~Palpanas, and Y.~Velegrakis.
\newblock Graph-query suggestions for knowledge graph exploration.
\newblock In {\em Proceedings of The Web Conference 2020}, pages 2549--2555,
  2020.

\bibitem{liu2011learning}
T.-Y. Liu.
\newblock {\em Learning to rank for information retrieval}.
\newblock Springer Science \& Business Media, 2011.

\bibitem{nsg3}
J.~Llad{\'o}s, E.~Mart{\'\i}, and J.~J. Villanueva.
\newblock Symbol recognition by error-tolerant subgraph matching between region
  adjacency graphs.
\newblock {\em IEEE Transactions on Pattern Analysis and Machine Intelligence},
  23(10):1137--1143, 2001.

\bibitem{lyzinski2016consistency}
V.~Lyzinski, K.~Levin, D.~E. Fishkind, and C.~E. Priebe.
\newblock On the consistency of the likelihood maximization vertex nomination
  scheme: Bridging the gap between maximum likelihood estimation and graph
  matching.
\newblock {\em Journal of Machine Learning Research}, 17(179):1--34, 2016.

\bibitem{lyzinski2017consistent}
V.~Lyzinski, K.~Levin, and C.~E. Priebe.
\newblock On consistent vertex nomination schemes.
\newblock {\em Journal of Machine Learning Research}, 20(69):1--39, 2019.

\bibitem{lyzinski2015community}
V.~Lyzinski, M.~Tang, A.~Athreya, Y.~Park, and C.~E. Priebe.
\newblock Community detection and classification in hierarchical stochastic
  blockmodels.
\newblock {\em IEEE Transactions on Network Science and Engineering},
  4(1):13--26, 2017.

\bibitem{marchette2011vertex}
D.~Marchette, C.~E. Priebe, and G.~Coppersmith.
\newblock Vertex nomination via attributed random dot product graphs.
\newblock In {\em Proceedings of the 57th ISI World Statistics Congress},
  volume~6, 2011.

\bibitem{bio3}
R.~Milo, S.~Shen-Orr, S.~Itzkovitz, .~Kashtan, D.~Chklovskii, and U.~Alon.
\newblock Network motifs: simple building blocks of complex networks.
\newblock {\em Science}, 298(5594):824--827, 2002.

\bibitem{bertozzi_2018}
Jacob~D. Moorman, Qinyi Chen, Thomas~K. Tu, Zachary~M. Boyd, and Andrea~L.
  Bertozzi.
\newblock Filtering methods for subgraph matching on multiplex networks.
\newblock {\em 2018 IEEE International Conference on Big Data (Big Data)},
  2018.

\bibitem{exemplar1}
D.~Mottin, M.~Lissandrini, Y.~Velegrakis, and T.~Palpanas.
\newblock Exemplar queries: a new way of searching.
\newblock {\em The VLDB Journal}, 25(6):741--765, 2016.

\bibitem{natarajan2013learning}
N.~Natarajan, I.~S. Dhillon, P.~K. Ravikumar, and A.~Tewari.
\newblock Learning with noisy labels.
\newblock In {\em Advances in neural information processing systems}, pages
  1196--1204, 2013.

\bibitem{newman2006modularity}
M.~E.~J. Newman.
\newblock Modularity and community structure in networks.
\newblock {\em Proceedings of the National Academy of Sciences},
  103(23):8577--8582, 2006.

\bibitem{park10:_dynam_bayes}
Y.~Park, C.~Moore, and J.~S. Bader.
\newblock Dynamic networks from hierarchical {B}ayesian graph clustering.
\newblock {\em PLOS ONE}, 5, 2010.

\bibitem{patsolic2017vertex}
H.~G. Patsolic, Y.~Park, V.~Lyzinski, and C.~E. Priebe.
\newblock Vertex nomination via local neighborhood matching.
\newblock {\em Statistical Analysis and Data Mining: The ASA Data Science
  Journal}, 13(3):229--244, 2020.

\bibitem{Peixoto_HSBM}
Tiago~P. Peixoto.
\newblock Hierarchical block structures and high-resolution model selection in
  large networks.
\newblock {\em Physical Review X}, 4(011047):1--18, 2014.

\bibitem{rastogi2017vertex}
P.~Rastogi, V.~Lyzinski, and B.~Van~Durme.
\newblock Vertex nomination on the cold start knowledge graph.
\newblock {\em Human Language Technology Center of Excellence: Technical
  report}, 2017.

\bibitem{rastogi2019neural}
P.~Rastogi, A.~Poliak, V.~Lyzinski, and B.~Van~Durme.
\newblock Neural variational entity set expansion for automatically populated
  knowledge graphs.
\newblock {\em Information Retrieval Journal}, 22(3-4):232--255, 2019.

\bibitem{sales-pardo07:_extrac}
M.~Sales-Pardo, R.~Guimer\`{a}, A.~A. Moreira, and L.~A.~N. Amaral.
\newblock Extracting the hierarchical organization of complex systems.
\newblock {\em Proc. Natl. Acad. Sci. U.S.A.}, 104, 2007.

\bibitem{sg2}
G.~M Slota and K.~Madduri.
\newblock Fast approximate subgraph counting and enumeration.
\newblock In {\em 2013 42nd International Conference on Parallel Processing},
  pages 210--219. IEEE, 2013.

\bibitem{sussman2018matched}
D.~L. Sussman, V.~Lyzinski, Y.~Park, and C.~E. Priebe.
\newblock Matched filters for noisy induced subgraph detection.
\newblock {\em IEEE Transactions on Pattern Analysis and Machine Intelligence},
  2019.

\bibitem{suwan2015bayesian}
S.~Suwan, D.~S. Lee, and C.~E. Priebe.
\newblock Bayesian vertex nomination using content and context.
\newblock {\em Wiley Interdisciplinary Reviews: Computational Statistics},
  7(6):400--416, 2015.

\bibitem{tang14:_nonpar}
M.~Tang, A.~Athreya, D.~L. Sussman, V.~Lyzinski, and C.~E. Priebe.
\newblock A nonparametric two-sample hypothesis testing for random dot product
  graphs.
\newblock {\em Bernoulli}, 23:1599--1630, 2017.

\bibitem{traag2019louvain}
V.~A. Traag, L.~Waltman, and N.~J. Van~Eck.
\newblock From louvain to leiden: guaranteeing well-connected communities.
\newblock {\em Scientific reports}, 9(1):1--12, 2019.

\bibitem{sg4}
J.~R. Ullmann.
\newblock An algorithm for subgraph isomorphism.
\newblock {\em Journal of the ACM (JACM)}, 23(1):31--42, 1976.

\bibitem{yoder2018vertex}
J.~Yoder, L.~Chen, H.~Pao, E.~Bridgeford, K.~Levin, D.~E. Fishkind, C.~E.
  Priebe, and V.~Lyzinski.
\newblock Vertex nomination: The canonical sampling and the extended spectral
  nomination schemes.
\newblock {\em Computational Statistics \& Data Analysis}, 145:106916, 2020.

\bibitem{zuo2014open}
X.-N. Zuo, J.~S. Anderson, P.~Bellec, R.~M. Birn, B.~B. Biswal, J.~Blautzik,
  J.~C.~S. Breitner, R.~L. Buckner, V.~D. Calhoun, F.~X. Castellanos, et~al.
\newblock An open science resource for establishing reliability and
  reproducibility in functional connectomics.
\newblock {\em Scientific data}, 1(1):1--13, 2014.

\end{thebibliography}

\end{document}